%% file: top.tex
\documentclass[final]{cvpr}
\usepackage{times}
\usepackage{epsfig}
\usepackage{graphicx}
\usepackage{amsmath}
\usepackage{amssymb}
\usepackage[ruled,vlined]{algorithm2e}
\usepackage{bbold}
\usepackage{float}
\usepackage{bbm}
\usepackage{booktabs} % for the table
\usepackage{subcaption}
\usepackage{verbatim}
\usepackage[pagebackref=true,breaklinks=true,colorlinks,bookmarks=false]{hyperref}
\usepackage{cleveref}

\def\cvprPaperID{1783} % *** Enter the CVPR Paper ID here
\def\confYear{CVPR 2021}

\begin{document}

%%%%%%%%% TITLE
% \title{3D Registration for Self-Occluded Objects in Context\\Supplementary Material}
\title{3D Registration for Self-Occluded Objects in Context}

\author{Zheng Dang\\
        Xi'an Jiaotong University\\
    	{\tt\small dangzheng713@stu.xjtu.edu.cn}\\
        % For a paper whose authors are all at the same institution,
        % omit the following lines up until the closing ``}''.
        % Additional authors and addresses can be added with ``\and'',
        % just like the second author.
        % To save space, use either the email address or home page, not both
        \and
    	Fei Wang\\
    	Xi'an Jiaotong University\\
    	{\tt\small wfx@mail.xjtu.edu.cn}\\
    	\and
    	Mathieu Salzmann\\
    	EPFL-CVLab \& ClearSpace\\
    	{\tt\small mathieu.salzmann@epfl.ch}
}

\maketitle
% \setkeys{Gin}{draft=True}
\input{tex/defs}
\input{tex/abstract}
\input{tex/introduction}

\input{tex/related_work}

\input{tex/methodology}
\input{tex/experiment}

\input{tex/conclusion}

{\small
\bibliographystyle{ieee_fullname}
\bibliography{bibtex/string,bibtex/vision}
}

\end{document}

%% file: tex/defs.tex
\newcommand{\bX}{\mathcal{X}}
\newcommand{\btX}{\tilde{\mathcal{X}}}
\newcommand{\bx}{\mathbf{x}}
\newcommand{\bbx}{\bar{\mathbf{x}}}
\newcommand{\bY}{\mathcal{Y}}
\newcommand{\by}{\mathbf{y}}
\newcommand{\bby}{\bar{\mathbf{y}}}
\newcommand{\bfx}{\mathbf{f}^x}
\newcommand{\bfy}{\mathbf{f}^y}
\newcommand{\bthx}{\theta^x}
\newcommand{\bthy}{\theta^y}
\newcommand{\bK}{\mathbf{K}}
\newcommand{\bT}{\mathcal{T}}
\newcommand{\bt}{\mathbf{t}}
\newcommand{\bS}{\mathcal{S}}
\newcommand{\bbS}{\mathcal{\bar{S}}}
\newcommand{\bP}{\mathcal{P}}
\newcommand{\bbP}{\mathcal{\bar{P}}}
\newcommand{\bM}{\mathcal{M}}
\newcommand{\bbM}{\mathcal{\bar{M}}}
\newcommand{\bA}{\mathcal{A}}
\newcommand{\bH}{\mathbf{H}}
\newcommand{\bR}{\mathbf{R}}
\newcommand{\bW}{\mathbf{W}}
\newcommand{\bU}{\mathbf{U}}
\newcommand{\bI}{\mathbf{I}}
\newcommand{\bV}{\mathbf{V}}
\newcommand{\ba}{\mathbf{a}}
\newcommand{\bb}{\mathbf{b}}
\newcommand{\bg}{\mathbf{g}}

\newcommand{\hR}{\hat{R}}
\newcommand{\hht}{\hat{\textbf{t}}}
\newcommand{\gR}{R_{gt}}
\newcommand{\gt}{\textbf{t}_{gt}}
\newcommand{\norm}[1]{\left\lVert#1\right\rVert}

\newcommand{\MS}[1]{\textcolor{red}{{\bf #1}}}
\newcommand{\ms}[1]{\textcolor{red}{#1}}
\newcommand{\ZD}[1]{{\color{blue}{\bf ZD: #1}}}
\newcommand{\zd}[1]{{\color{blue}{#1}}}

%% file: tex/abstract.tex
\begin{abstract}
    \vspace{-.2cm}
    While much progress has been made on the task of 3D point cloud registration, there still exists no learning-based method able to estimate the 6D pose of an object observed by a 2.5D sensor in a scene. The challenges of this scenario include the fact that most measurements are outliers depicting the object's surrounding context, and the mismatch between the complete 3D object model and its self-occluded observations.
    We introduce the first deep learning framework capable of effectively handling this scenario. Our method consists of an instance segmentation module followed by a pose estimation one. It allows us to perform 3D registration in a one-shot manner, without requiring an expensive iterative procedure. We further develop an on-the-fly rendering-based training strategy that is both time- and memory-efficient. Our experiments evidence the superiority of our approach over the state-of-the-art traditional and learning-based 3D registration methods.
\end{abstract}

%% file: tex/introduction.tex
\vspace{-.6cm}
\section{Introduction}
\input{fig/teaser}

3D registration aims to determine the rigid transformation, i.e., 3D rotation and 3D translation, between two 3D point sets. The traditional approach to this problem is the Iterative Closest Point (ICP) algorithm~\cite{Besl92}. In its vanilla version, this iterative algorithm easily gets trapped in poor local optima. While globally-optimal solutions~\cite{Yang15,Zhou16} have been proposed to remedy this, they 
%come at a high computational cost and 
lack robustness to noise, thus greatly reducing their practical applicability.

Recently, learning-based methods~\cite{Aoki19,Wang19e} were shown to outperform the previous traditional, optimization-based techniques.
%in terms of both speed and robustness to noise. In particular, DCP~\cite{Wang19e} provides a one-shot strategy that prevents the need for an expensive iterative process. Unfortunately, these methods cannot handle the partial-to-partial registration scenario, thus only
While these initial methods were designed to work under the unrealistic assumption that both point sets are fully observed, even at test time, follow-up works focused on the more practical partial-to-partial registration scenario~\cite{Wang19f,Yew20}. %\MS{Is it reasonable to group PRNet, PRMNet and DGR~\cite{Choy20} in this statement?}
%This was addressed by PRNet~\cite{Wang19f}, but by falling back to an iterative keypoint-matching procedure, relying on a reinforcement learning-inspired actor-critic strategy. This drastically increased the complexity of the resulting network, requiring four 32GB GPUs for training, instead of just one for DCP. 
Nevertheless, these techniques still assume the observations to depict a point-cloud in full 3D. 
%\MS{Correct?}
%\ZD{DGR has an experiment on the KITTI outdoor LiDAR scan dataset. But it doesn't have the experiment about the object.}
In practice, however, many sensors, such as depth cameras and LiDARs, only provide 2.5D measurements. Furthermore, when focusing on objects, as we do, and not entire scenes, as in, e.g.,~\cite{Choy20}, one needs to handle the measurements coming from the object's context, i.e., the scene itself, which are irrelevant to estimate the object 6D pose but are nonetheless captured by the sensor.

Only few methods have attempted to estimate the 6D pose of an object in a scene from depth data~\cite{Vidal18,Yang19,Yang20}. 
%In particular, the most recent method, TEASER++~\cite{Yang20}, handles the presence of contextual information by casting the problem as an outlier removal task. 
To this end, these methods extract handcrafted representations of the 3D points, such as Point Pair Features (PPF)~\cite{Vidal18} or Fast Point Feature Histograms (FPFH)~\cite{Yang19,Yang20} %\MS{Check and define these acronyms.}, 
and handle the presence of contextual information by casting the problem as an outlier removal task. %\MS{Is this also true for Vidal18?}\ZD{I think it's true. Vidal18 uses a process which is quite close to RANSAC.}
As such, they do not truly focus on the 2.5D nature of the measurements, and, for example, the effectiveness of the most recent one, TEASER++~\cite{Yang20},
%Because the focus of~\cite{Yang19,Yang20} was not the 2.5D nature of the measurements, but rather to remove the outliers corresponding to the object's context, the effectiveness of TEASER++ 
was only demonstrated in a restricted scenario, where the reference object point-cloud was in one-to-one correspondence with that observed in the scene context. %In other words, while effective at handling large amounts of outliers, TEASER++ does not account for the mismatch between the object model, which is typically a complete 3D model, and the 2.5D sensor measurements in which the object is self-occluded.
Furthermore, these methods all rely on handcrafted features, and, to the best of our knowledge, 6D pose estimation of a self-occluded object in context, as illustrated by Fig.~\ref{fig:teaser}, has never been tackled in a deep learning fashion.

In this paper, we address this by introducing a deep learning framework for 3D registration of self-occluded objects in context. Our approach consists of an instance segmentation module to separate the object from the surrounding scene, followed by a 6D pose estimation module able to handle the mismatch between the complete 3D model and the observed, self-occluded point-set. Our method 
%is end-to-end trainable\ZD{We have to train YOLACT and the pose estimation network separately, could it still be called end-to-end?} and 
works in a one-shot fashion; at test time, we estimate the object pose by a single forward pass through the network, without the need for an expensive iterative procedure. This makes our approach amenable to practical applications relying on 2.5D sensors and requiring real-time performance, such as autonomous navigation and robotics manipulation.

Our contributions can be summarized as follows:
\vspace{-1.5mm}
\begin{itemize}
    \itemsep-0.2em 
    \item We tackle, for the first time, the challenging problem of estimating the 6D pose of a self-occluded object in context in a learning-based manner.
    \item We develop a one-shot framework to address this task, consisting of an instance segmentation and a pose estimation module.
    \item We design a memory- and computation-effective training method based on on-the-fly rendering, which prevents the need for expensive offline rendering and storing of the generated data, as required by standard procedures, such as BlenderProc~\cite{Denninger19}, while giving us access to virtually infinite amounts of training data.
\end{itemize}
\vspace{-1mm}
We demonstrate the effectiveness of our approach using the ModelNet40 benchmark, from which we generate depth maps to match our application scenario. Our approach significantly outperforms the state-of-the-art traditional and leaning-based registration techniques~\cite{Wang19e,Wang19f}, including our main competitor TEASER++~\cite{Yang19,Yang20}. We will make our code publicly available.
%\MS{Can we show results on real depth maps? This would make the paper much more convincing, even if they are just qualitative.}\ZD{Have some preliminary result, cleaning up the pipeline and the result.}

%% file: fig/teaser.tex
\begin{figure}[!ht]
    \centering
    \begin{subfigure}[b]{0.47\textwidth}
    \centering
    \includegraphics[width=4cm]{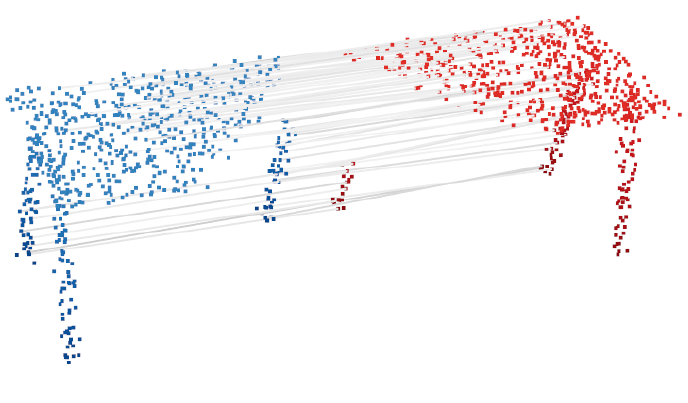}
    \includegraphics[width=4cm]{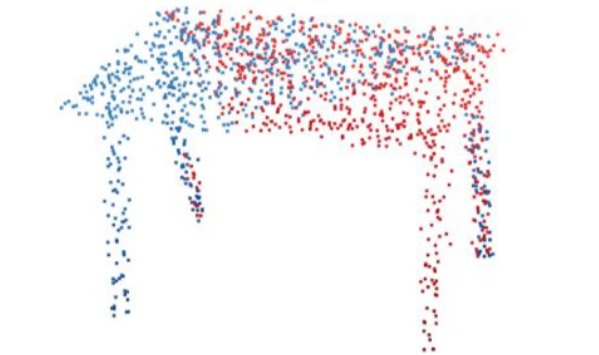}
    \caption{RPMNet}
    \end{subfigure}
    
    \begin{subfigure}[b]{0.47\textwidth}
    \centering
    \includegraphics[width=3.2cm]{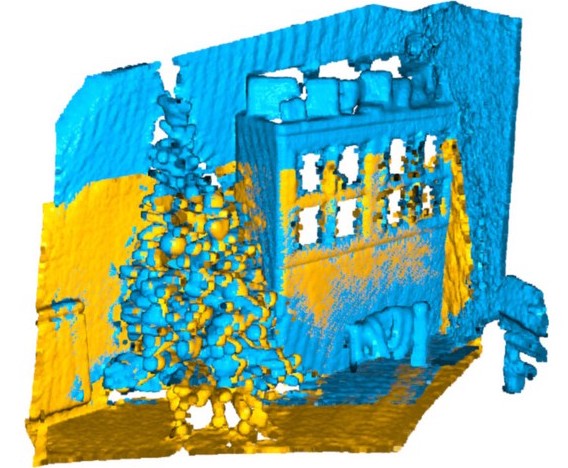}
    \includegraphics[width=4.8cm]{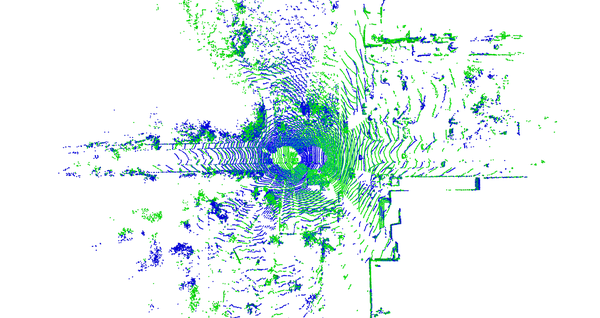}
    \caption{DGR}
    \end{subfigure}

    \begin{subfigure}[b]{0.47\textwidth}
    \centering
    \includegraphics[width=4cm]{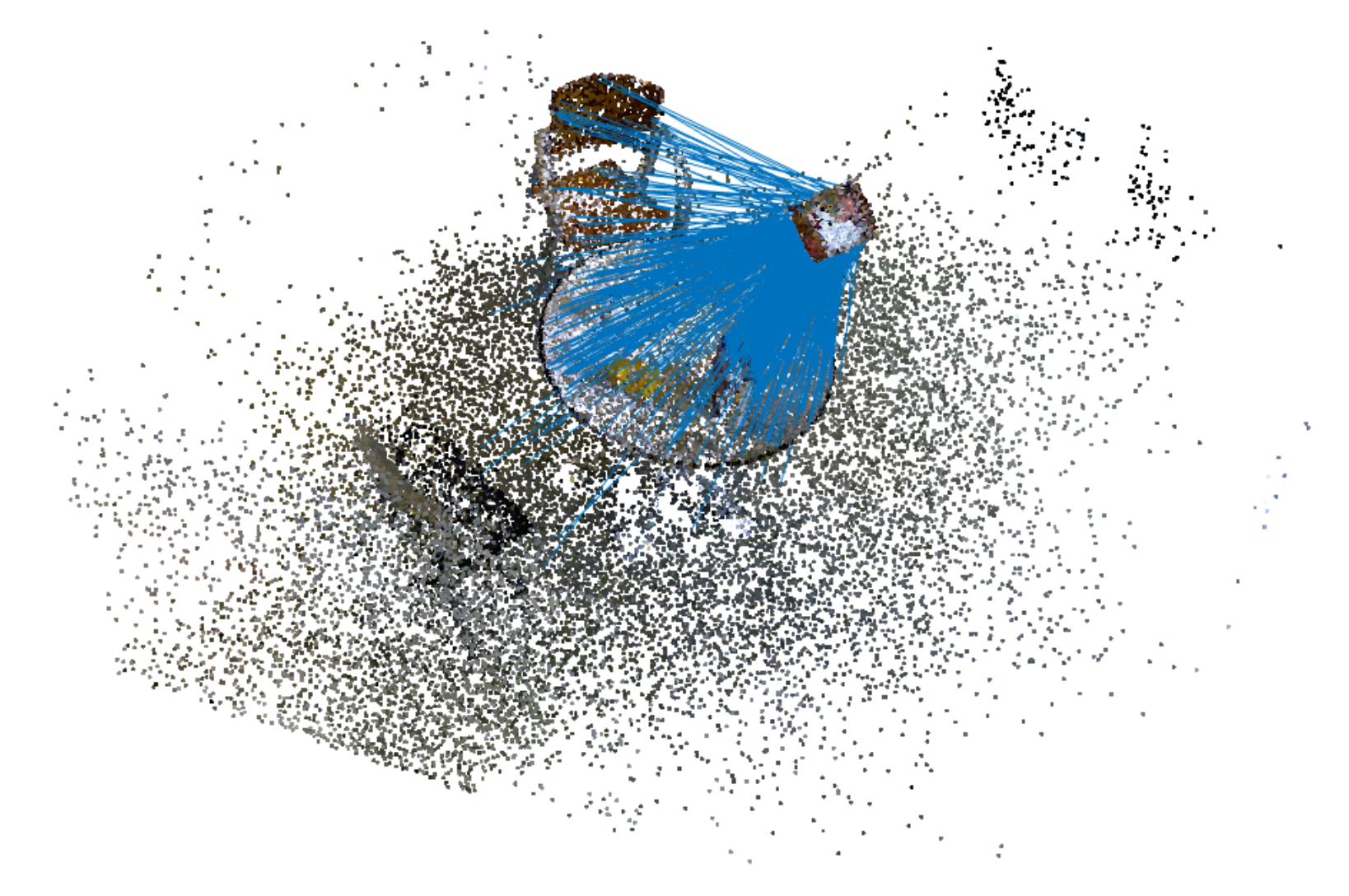}
    \includegraphics[width=4cm]{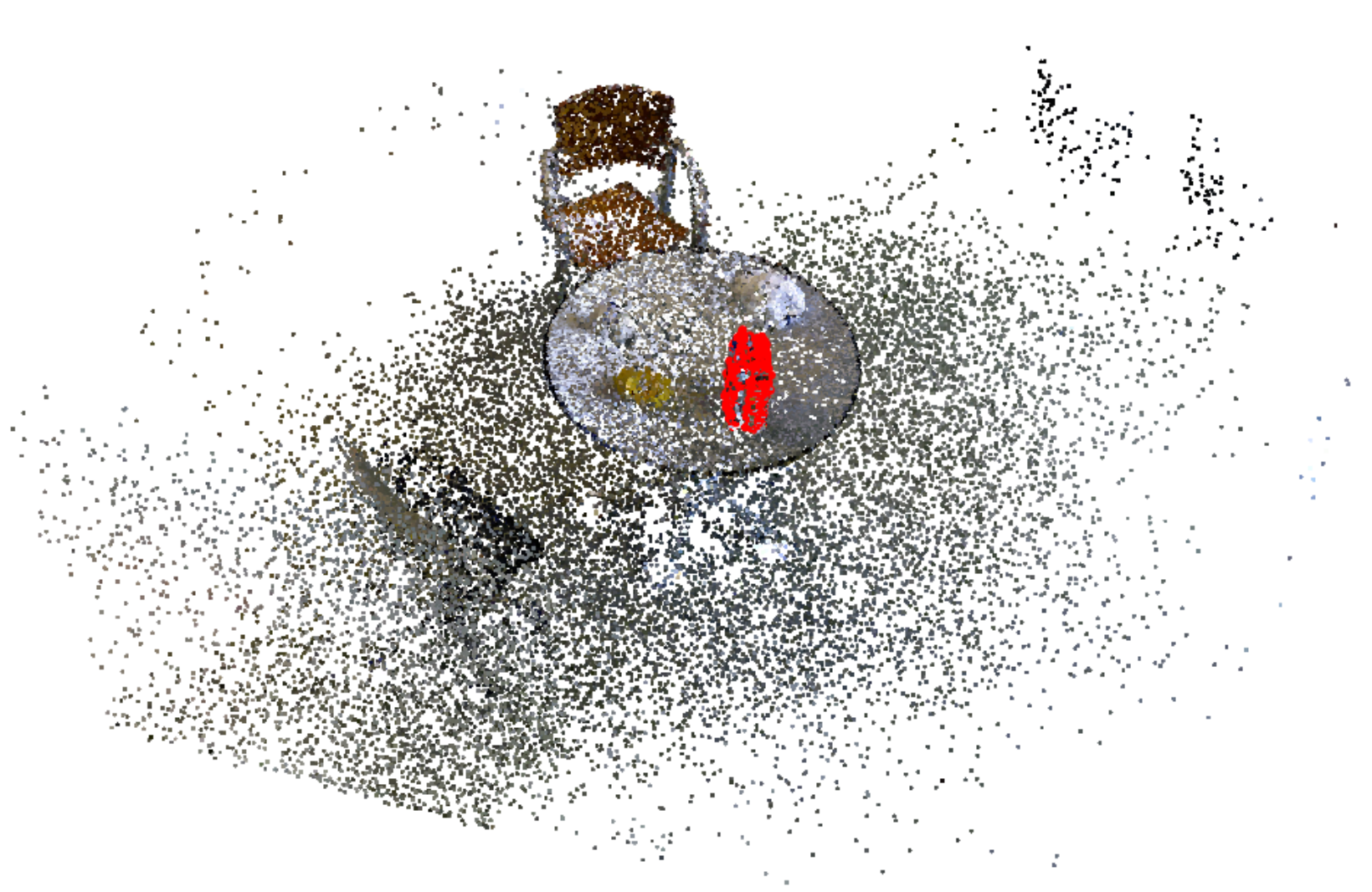}
    \caption{TEASER++}
    \end{subfigure}
        
    \begin{subfigure}[b]{0.47\textwidth}
    \centering
    \includegraphics[width=3.6cm]{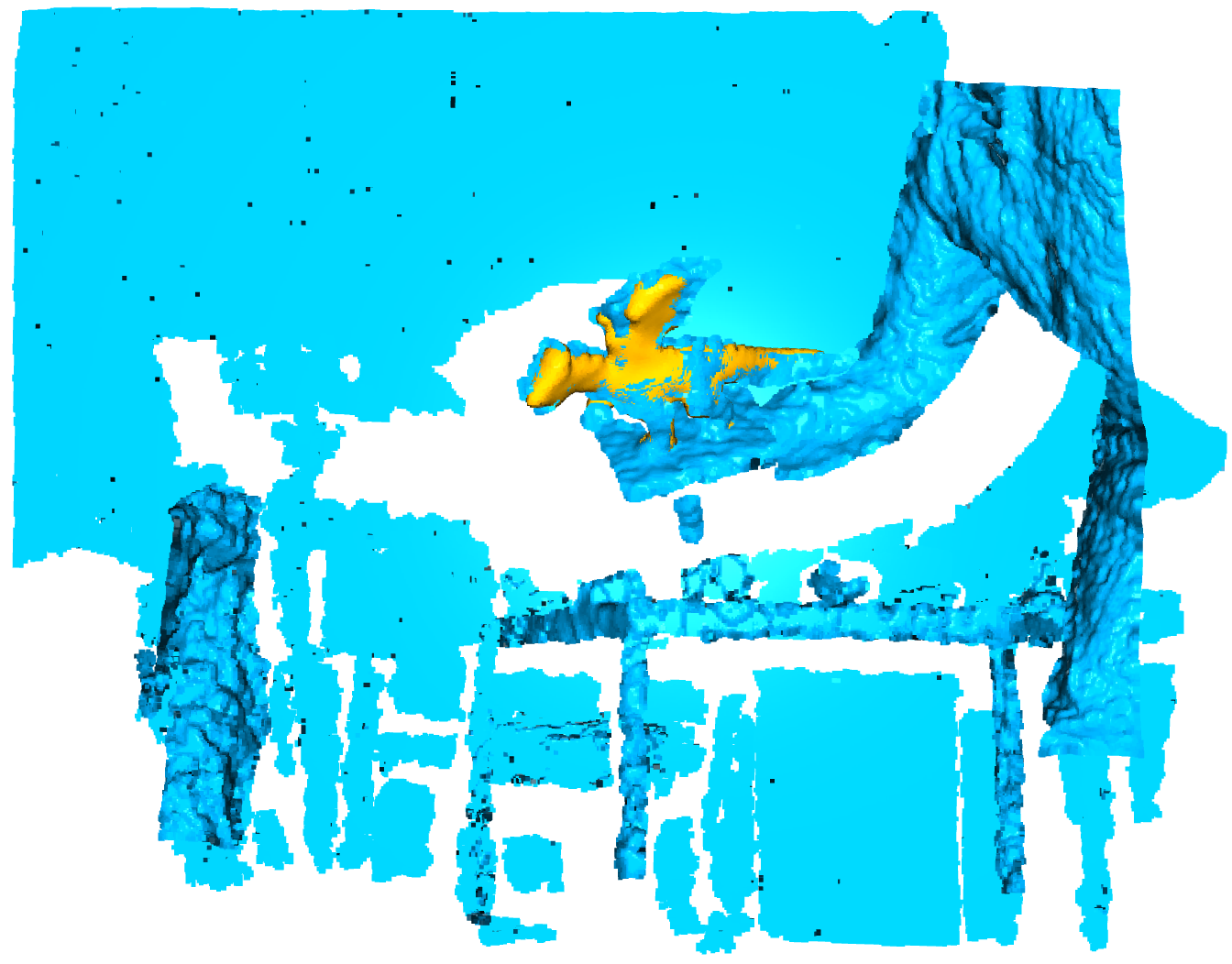}
    \hspace{0.3cm}
    \includegraphics[width=3.8cm]{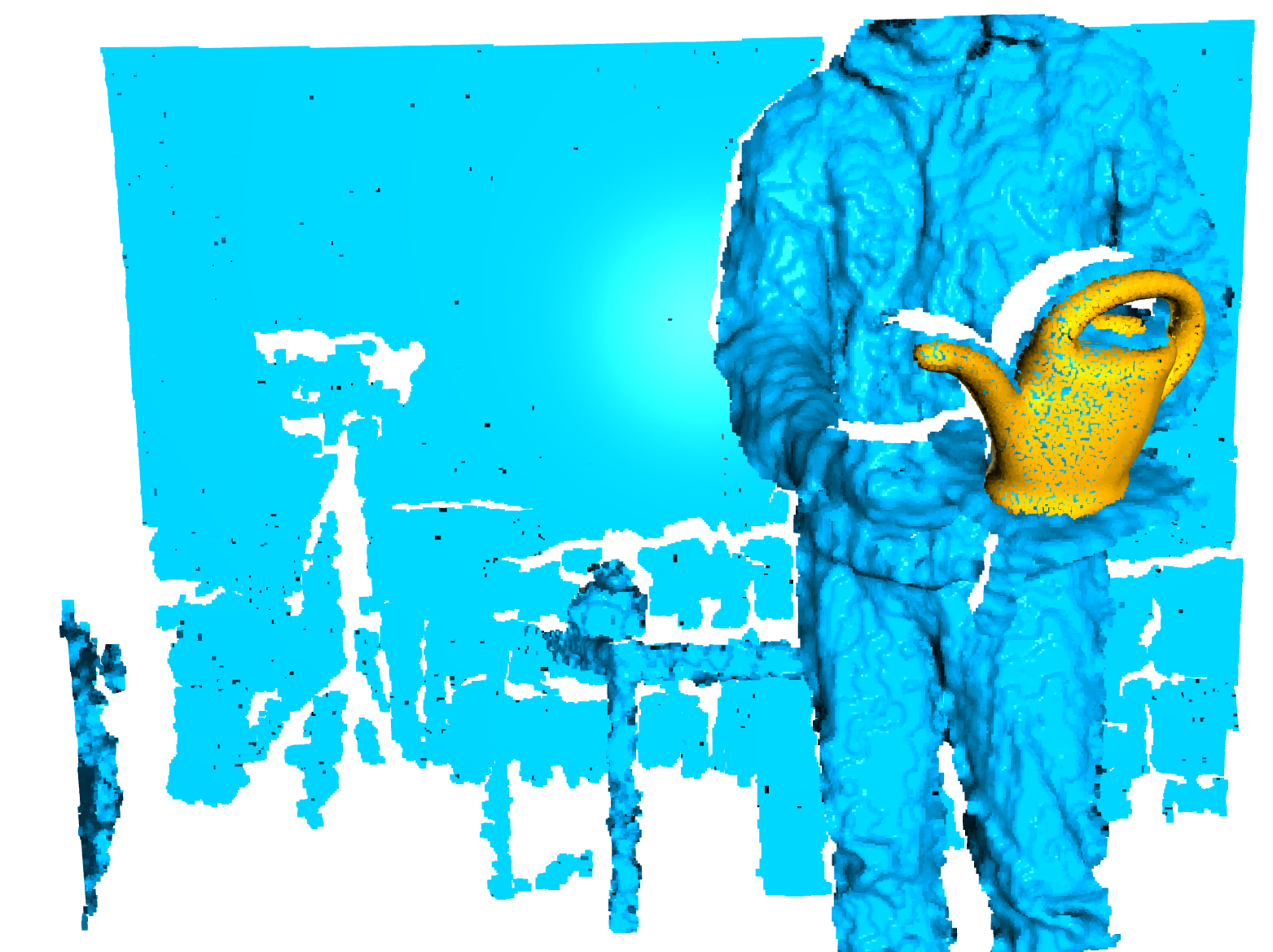}
    \caption{Ours}
    \end{subfigure}
    \setlength{\belowcaptionskip}{-25pt}
    \setlength{\abovecaptionskip}{-5pt}
    \caption{\label{fig:teaser} 
    {\small In contrast to object-to-object registration, as done by, e.g., RPMNet~\cite{Yew20}, and to scene-to-scene registration, as performed by, e.g., DGR~\cite{Choy20}, we tackle the challenging task of estimating the 6D pose of an object in a scene captured with a 2.5D sensor. While TEASER++~\cite{Yang20} showcased this for the case where the object model is in one-to-one correspondence with the observed point set, we propose a learning-based approach able to handle the scenario where the model is self-occluded. (RPMNet, DGR and TEASER++ were borrowed from the original papers.)
    }}
    % Different from partial-to-partial matching in RPMNet and the registration of two scenes in DGR, only TEASER++ and ours method focus on the registration of the object in a scene. However, TEASER++ assumes that the model is the point set observed in the scene, whereas we consider a complete model. (Figures of RPMNet, DGR and TEASER++ are borrowed from its original paper.)
\end{figure}

%% file: tex/related_work.tex
\section{Related Work}

\textbf{Traditional point cloud registration.} 
ICP is the best-known algorithm for solving the point cloud registration problem. It comprises two steps: One whose goal is to find the closest target point for each source point to generate 3D-3D correspondences, and the other that computes the rigid transformation from these correspondences by solving a least-square problem. These two steps are repeated until a termination condition is satisfied. Several variants, such as Generalized-ICP~\cite{Segal09} and Sparse ICP~\cite{Bouaziz13}, have been proposed to improve  robustness to noise and mismatches, and we refer the reader to~\cite{Pomerleau15,Rusinkiewicz01} for a complete review of ICP-based strategies. The main drawback of these methods is their requirement for a reasonable initialization to converge to a good solution. Only relatively recently has this weakness been addressed by the globally-optimal registration method Go-ICP~\cite{Yang15}. In essence, this approach follows a branch-and-bound strategy to search the entire 3D motion space $SE(3)$. Motivated by this, other approaches to finding a global solution have been proposed, via, e.g., Riemannian optimization~\cite{Rosen19}, convex relaxation~\cite{Maron16}, and mixed-integer programming~\cite{Jzatt20}. While globally optimal, these methods all come at a much higher computational cost than vanilla ICP. This was, to some degree, addressed by the Fast Global Registration (FGR) algorithm~\cite{Zhou16}, which leverages a local refinement strategy to speed up computation. While effective, FGR still suffers from the presence of noise and outliers in the point sets, particularly because, as vanilla ICP, it simply relies on 3D point-to-point distance to establish correspondences. In principle, this can be addressed by designing point descriptors that can be more robustly matched. Over the years, several works have tackled this task, in both a non learning-based~\cite{Johnson99,Rusu08,Rusu09} and learning-based~\cite{Zeng17b,Khoury17} fashion. Nowadays, however, these approaches are outperformed by end-to-end learning frameworks, which directly take the point sets as input.

\textbf{End-to-end learning with point sets.} 
A key requirement to enable end-to-end learning-based registration was the design of deep networks acting on unstructured sets. Deep sets~\cite{Zaheer17} and PointNet~\cite{Qi17} constitute the pioneering works in this direction. They use shared multilayer perceptrons to extract high-dimensional features from the input point coordinates, and exploit a symmetric function to aggregate these features. This idea was then extended in PointNet++~\cite{Qi17a} via a modified  sampling strategy to robustify the network to point clouds of varying density, in DGCNN~\cite{Wang18b} by building a graph over the point cloud, in PointCNN~\cite{Li18c} by learning a transformation of the data so as to be able to process it with standard convolutional layers, and in PCNN~\cite{Atzmon18} via an additional  extension operator before applying the convolutions. While the above-mentioned works focused on other tasks than us, such as point cloud classification or segmentation, end-to-end learning for registration has recently attracted a growing attention. In particular, PointNetLK~\cite{Aoki19} combines the PointNet backbone with the traditional, iterative Lucas-Kanade (LK) algorithm~\cite{Lucas81} so as to form an end-to-end registration network; DCP~\cite{Wang19e} exploits DGCNN backbones followed by Transformers~\cite{Vaswani17} to establish 3D-3D correspondences, which are then passed through an SVD layer to obtain the final rigid transformation. While effective, PointNetLK and DCP cannot tackle the partial-to-partial registration scenario. That is, they assume that both point sets are fully observed, during both training and test time. This was addressed by PRNet~\cite{Wang19f} via a deep network designed to extract keypoints from each input set and match these keypoints. This network is then applied in an iterative manner, so as to increasingly refine the resulting transformation. 
% \MS{We need to discuss RPM-Net here.}
Similarly, RPM-Net~\cite{Yew20} builds on DCP, replacing its softmax layer with an optimal transport ones so as to handle outliers, and as PRNet, relies on an iterative strategy to refine the computed transformation.
In any event, the methods discussed above were designed to handle point-clouds in full 3D, and were thus not demonstrated for registration from 2.5D measurements. By contrast, DGR~\cite{Choy20}, which 
%\MS{Brief description of what DGR does}
uses a deep network to reject the outliers from an input set of correspondences, was shown to be applicable to depth data. Nevertheless, this was achieved in the context of registering two partial views of a scene, whereas we focus on estimating the 6D pose of an object captured in a scene.

To the best of our knowledge, only few methods have been proposed to address this challenging scenario. In particular,~\cite{Vidal18} relies on generating pose hypotheses via feature matching, followed by a RANSAC-inspired method to choose the candidate pose with the largest number of support matches.
%\MS{Brief description of Vidal18.}. %\MS{Brief description of TEASER++.} 
Similarly, TEASER~\cite{Yang19} and its improved version TEASER++~\cite{Yang20} take putative correspondences obtained via feature matching as input and remove the outlier ones by an adaptive voting scheme.
%is a revisited version which speed up the large scale SDP (semideﬁnite programming) computation, thus could be consider as a faster version of TEASER.
By relying on handcrafted features designed for 3D point-clouds, these methods do not explicitly address the case of 2.5D measurements. This is what we address in this work, and, motivated by the great progress of deep networks to address the classical 3D registration scenario, introduce the first deep learning framework capable of estimating the 6D pose of a 3D object observed with a 2.5D sensor in a scene.

%% file: tex/methodology.tex
\input{fig/pipeline}

\section{Methodology}

Let us now introduce our approach to estimating the 6D pose of an object, represented with a reference point cloud in full 3D, from 2.5D observations of this object in a scene. To address this challenging scenario, we develop the deep learning framework depicted by Fig.~\ref{fig:pipeline}. It relies on two main modules, an instance segmentation network and a pose estimation network, which we discuss in detail below.

\subsection{Instance Segmentation}
To tackle the realistic scenario where the object is immersed in a scene, we first process the input depth map with an instance segmentation module. Specifically, we build on the YOLACT~\cite{Bolya19} framework that has proven highly effective for image-based instance segmentation. To exploit this framework in our scenario, we replicate the depth map $I_{depth}$ three times, so as to obtain 3 channels, as an RGB image, which allows us to benefit from the pre-training of the model on an image dataset. Furthermore, instead of predicting multiple object categories, we train our model to discriminate between object and background, which we found to be better suited to handle depth that does not carry appearance information. Note that we nonetheless train our model using multiple object categories. Finally, after non-maximum suppression, we only keep the bounding box with the highest score, which we observed to work well in practice, as will be shown by our experiments, and as evidenced qualitatively in Fig.~\ref{synthetic_scene}. %\MS{Not sure if we want to show this here, or wait until the experiments section...}\ZD{For the reason it's a qualitative figure, I think here would be okay.}
Ultimately, our instance segmentation module produces a mask image, expressed as
\begin{equation}
    I_{mask} = \phi_{seg}(I_{depth})\;.
\end{equation}
We then use $I_{mask}$ to isolate the target point cloud, which we input to the pose estimation network described below.

\input{fig/pose_net}
\input{fig/synthetic_scene}
\vspace{-.2cm}
\subsection{Pose Estimation}
Let us now turn to the task of estimating the 6D pose of the object of interest. To this end, let $\bX \in \mathbb{R}^{M \times 3}$ and $\bY \in \mathbb{R}^{N \times 3}$ be two sets of 3D points sampled from the same object surface. We typically refer to $\bX$ as the source point set and to $\bY$ as the target point set. We obtain the source point set $\bX$ by uniform sampling from the mesh model, and target one $\bY$ from $I_{depth}$ and $I_{mask}$, assuming known camera intrinsic parameters. Registration then aims to find a rigid transformation $\bT$ that aligns $\bX$ to $\bY$. In this work, we focus on the case where the object is self-occluded. To be specific, we assume that each point in $\bY$ has a corresponding point in $\bX$, but not the opposite.

We build our pose estimation module on DCP~\cite{Wang19e}, extending it to handle the partial correspondence scenario. As illustrated in Fig.~\ref{pose_net}, we rely on the DCP-v2 design, which consists of a DGCNN~\cite{Wang18b} followed by a Transformer~\cite{Vaswani17}.
Specifically, the DGCNN takes a point set as input, constructs a k-NN graph from it, and then extracts point-wise features via standard convolutions on this graph, encoding diverse levels of context by max-pooling the local features and concatenating the resulting representations to the point-wise ones.
Let $\bthx$, resp. $\bthy$, be the final feature matrix, i.e., one $P$-dimensional feature vector per 3D point, for $\bX$, resp. $\bY$. The transformer then learns a function $\phi : \mathbb{R}^{M\times P} \times \mathbb{R}^{N\times P} \rightarrow \mathbb{R}^{M\times P}$, that combines the information of the two point sets. Ultimately, this produces descriptor matrices $\bfx$, resp. $\bfy$, for $\bX$, resp. $\bY$, written as
\begin{equation}
    \bfx = \bthx + \phi(\bthx, \bthy), \;\;\;
    \bfy = \bthy + \phi(\bthy, \bthx)\;.
\end{equation}
    
Given these matrices, we then form a score map $\bS \in \mathbb{R}^{M\times N}$ by computing the similarity between each source-target pair of descriptors. That is, we compute the $(i,j)$-th element of $\bS$ as
\begin{equation}
    \bS_{i,j} = <\bfx_i, \bfy_j>, \;\;\forall (i, j) \in [1,M]\times [1,N]\;,
\end{equation}
where $<\cdot, \cdot>$ is the inner product, and $\bfx_i, \bfy_j \in \mathbb{R}^P$. 

In DCP~\cite{Wang19e}, this score map is passed through a row-wise softmax so as to obtain correspondences. These correspondences are then processed within the network via an SVD layer to solve the Procrustes problem, and the resulting rigid transformation is compared to the ground-truth one with a mean squared error (MSE) loss. Here, when $I_{mask}$ can be assumed to be clean, we keep the softmax because each point in $\bY$ should be visible in $\bX$. However, we remove the SVD layer, which, if not carefully designed, can make the training process unstable~\cite{Dang18,Dang20,Wang19g,Yew20}. Instead, we impose direct supervision on the output of the softmax. To this end, let $\bM \in \{0, 1\}^{M \times N}$ be the matrix of ground-truth correspondences, with a $1$ indicating a correspondence between a pair of points. Such correspondences can be estimated using the ground-truth transformation matrix, as discussed in more detail in Section~\ref{no_background}. We then express our loss function as the negative log-likelihood
\begin{equation}
    \mathcal{L}(\bP, \bM) = \frac{- \sum\limits_{i =1}^{M}\sum\limits_{j = 1}^{N} (\log \bP_{i,j})\bM_{i, j}}{\sum\limits_{i = 1}^{M}\sum\limits_{j = 1}^{N} \bM_{i, j}}\;,
    \label{eq:nll}
\end{equation}
where $\bP$ is the output of the softmax, and where the denominator normalizes the loss value so that different training samples containing different number of correspondences have the same influence in the overall empirical risk.

\vspace{-.4cm}
\paragraph{Dealing with noisy target points.}
\input{table/algorithm}
Although the softmax works well when the points in $\bY$ all have a corresponding point in $\bX$, in practice, $\bY$ may include noise, due, for example, to an imperfect $I_{mask}$ prediction.
%Which is also the reason that SoftMax cannot be trained on BlenderProc dataset in Section~\ref{blenderproc_dataset}. 
To address this, we replace the softmax with an optimal transport layer~\cite{Yew20,Sarlin19} %\MS{I suggest we also cite our arXiv paper.} \MS{Maybe not, in fact.}
including an outlier bin to handle the noise.
    
Specifically, we extend the score matrix $\bS$ by one row and one column to form an augmented score matrix $\bbS$. The values at the newly-created positions in $\bbS$ are set to
\begin{equation}
    \bbS_{i, N + 1} = \bbS_{M + 1, j} = \bbS_{M + 1, N + 1} = \alpha,
\end{equation}
$\forall i \in [1, M], \;\forall j \in [1, N]$, where $\alpha \in \mathbb{R}$ is a fixed parameter, which set to be $0.01$ in practice. The values at the other indices directly come from $\bS$. Given the augmented score map $\bbS$, we aim to find a partial assignment $\bbP \in \mathbb{R}^{(M+1)\times (N+1)}$, defining correspondences between the two point sets, extended with the outlier bins. Let $\bU (\ba, \bb)$ be the set of probability matrices defined as
\begin{equation}
    \left\{\bbP \in \mathbb{R}_{+}^{(M+1)\times (N+1)}: \bbP \mathbbm{1}_{M+1} = \ba \: \text{and}\: \bbP^{\top} \mathbbm{1}_{N+1} = \bb \right\}\;,
\end{equation}
where $\ba = [\mathbbm{1}_{M}^{\top}, N]^{\top}$, and $\bb = [\mathbbm{1}_{N}^{\top}, M]^{\top}$, with $\mathbbm{1}_{M} = [1, 1, ..., 1]^{\top}\in \mathbb{R}^{M}$. Then, from the optimal transport theory~\cite{Peyre19,Cuturi13}, an assignment probability matrix $\bbP$ can be obtained by solving
\begin{equation}
    \min_{\bbP \in \bU (\ba, \bb)} \langle \bbS, \bbP \rangle - \lambda E(\bbP)\;,
\label{entropy_func}
\end{equation}
where $\langle\cdot,\cdot\rangle$ is the Frobenius dot product and $E(\cdot)$ is an entropy regularization term defined as $E(\bbP) = - \displaystyle\sum_{i, j} \bbP_{i, j} \left(\log(\bbP_{i, j}) - 1\right)$. In practice, this optimization problem can be solved by using the log-domain Sinkhorn algorithm, which we summarized in Algorithm~\ref{alg:sinkhorn}, where the matrix operator $logsumexp(\bA) = log(exp(\bA_{1,1}) + ... + exp(\bA_{i,j}) + ... + exp(\bA_{M,N}))$. Since all operations performed by this algorithm are differentiable, the training errors can be backpropagated to the rest of the network. We then still use the negative log-likelihood of Eq.~\ref{eq:nll} as loss function, but replacing $\bM$ with $\bbM$, which contains an extra row and column acting as outlier bins.
    
\subsection{Training via On-the-fly Rendering}
\label{sec:training}

To train our network, we follow a two-stage procedure, starting with the instance segmentation module and following with the pose estimation one. 
%\MS{Why don't we train end-to-end?} \MS{Why don't we use the on-the-fly approach for the segmentation module as well?}\ZD{Main reason is time. By the time of deadline we don't have enough time to fully revisited our method... We could try these ideas after the deadline.} 
Unfortunately, there exist virtually no real dataset providing high-quality depth data for 6D object pose estimation in context. 
%\MS{Can we explain why our approach does not apply to lower quality datasets, such as BOP?}
For instance, the datasets in the BOP challenge~\cite{Hodan18} mostly provide relatively low-quality depth maps, with very sparse  depth measurements for the object. Such low-quality depth observations, however, are not representative of the resolutions that modern sensors can now achieve~\cite{Lu20}.
%, making instance segmentation virtu which is incomplete could highly influence the performance of our instance segmentation network.} 
%In this section, we discuss our approach to addressing this by generating high-quality synthetic data.
To better reflect the progress of these sensors, in this work, we rely on BlenderProc~\cite{Denninger19} to generate high-quality synthetic data, as will be discussed in more detail in Section~\ref{with_background}. However, even such synthetic data remains imperfect at the object boundaries, mixing the object depth with the background one, as shown in Fig.~\ref{fig:syn_qualitative}. Such outliers at the boundaries affect the training process, and we therefore use BlenderProc to train our instance segmentation module but not our pose estimation one. Instead, and to simultaneously benefit from having access to virtually infinite amounts of training data, we follow an on-the-fly rendering strategy. Specifically, we use Pytorch3D~\cite{ravi20} to generate scenes that contain individual objects without background, thus fulfilling the pose estimation network's assumption and precluding the presence of outliers at the object boundaries. The use of Pytorch3D allows us to interface the rendering process with the network training, thus preventing the need to generate and store a fixed set of training samples in a pre-processing stage.

%% file: fig/pipeline.tex
\begin{figure}[t]
    \centering
    \setlength{\belowcaptionskip}{-1.5cm}
    \includegraphics[width=8cm]{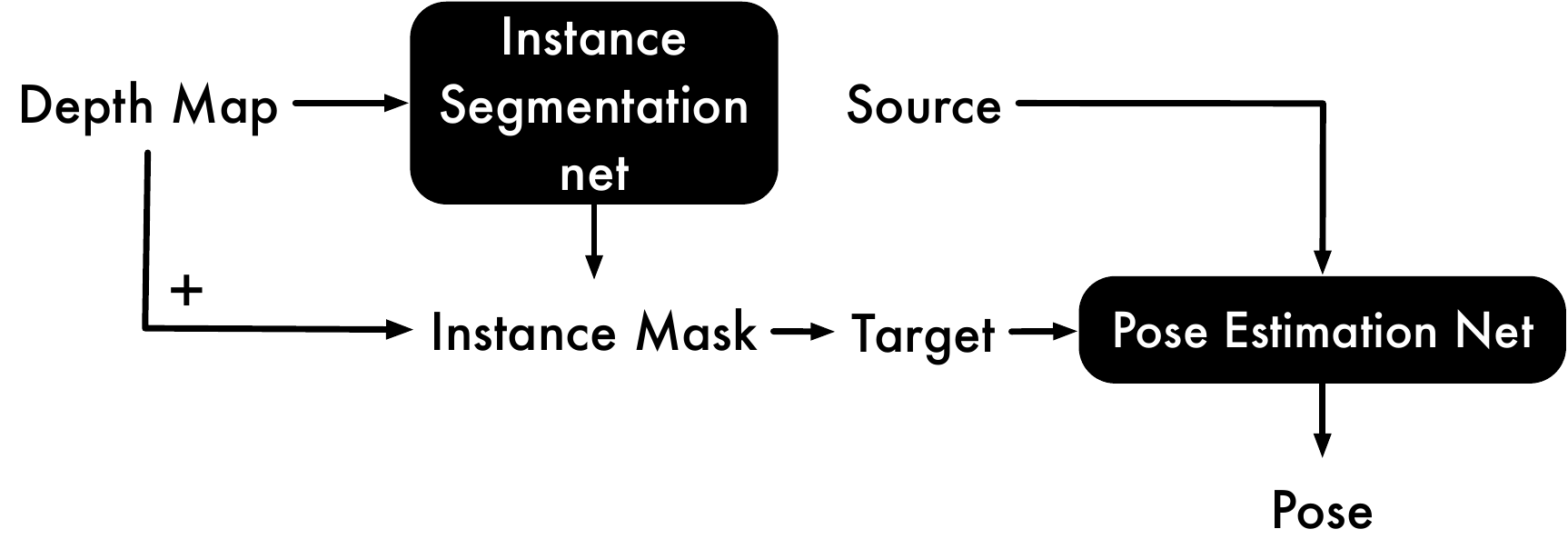}
    \caption{\label{pipe} Illustration of our framework.}%\MS{Should we define a name for our method?}}
    \label{fig:pipeline}
\end{figure}

%% file: fig/pose_net.tex
\begin{figure}[t]
    \centering
    \setlength{\belowcaptionskip}{-10pt}
    \includegraphics[width=8cm]{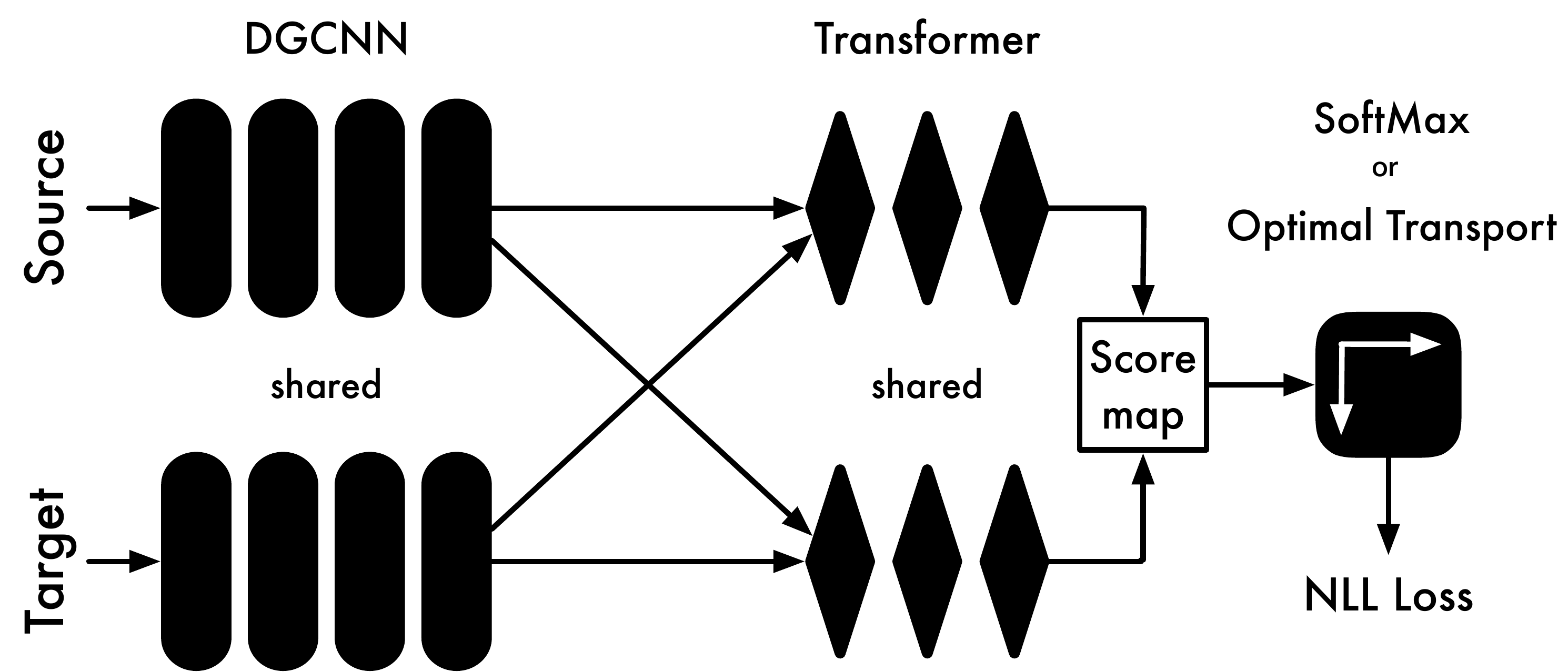}
    \caption{\label{pose_net} Architecture of the pose estimation network.}
\end{figure}

%% file: fig/synthetic_scene.tex
\begin{figure*}[t]
    \setlength{\belowcaptionskip}{-10pt}
    \includegraphics[width=3.4cm]{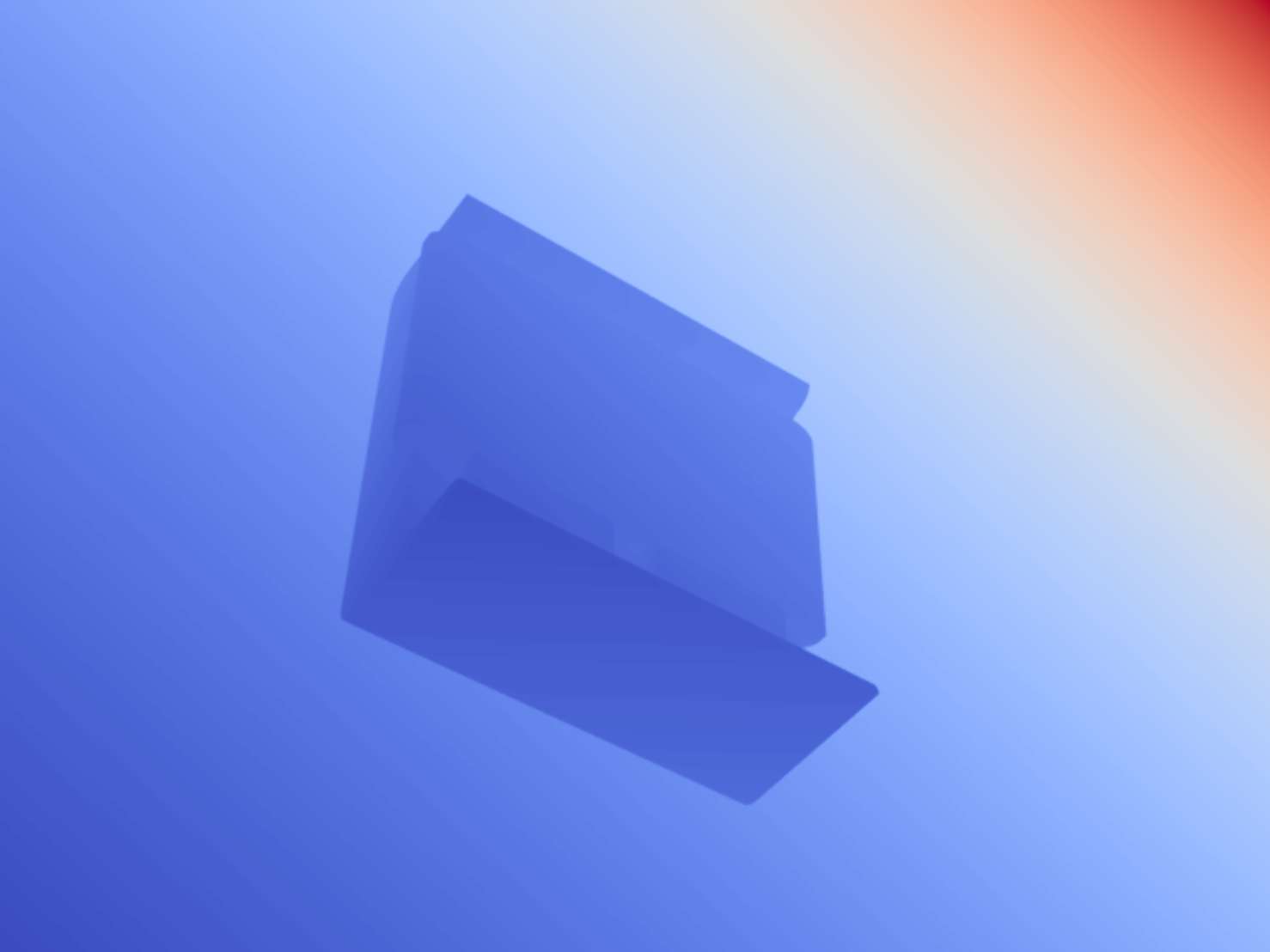}
    \includegraphics[width=3.4cm]{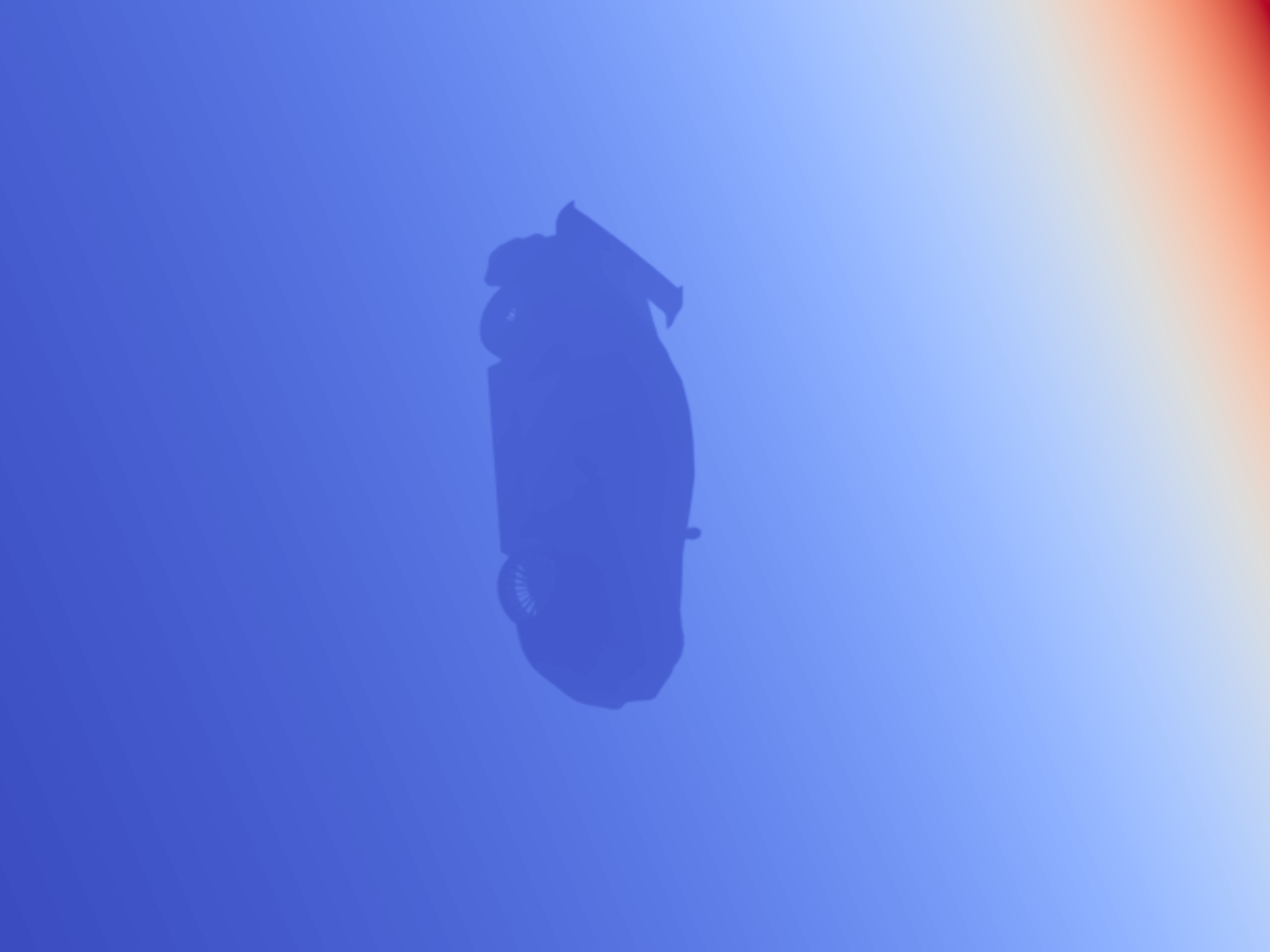}
    \includegraphics[width=3.4cm]{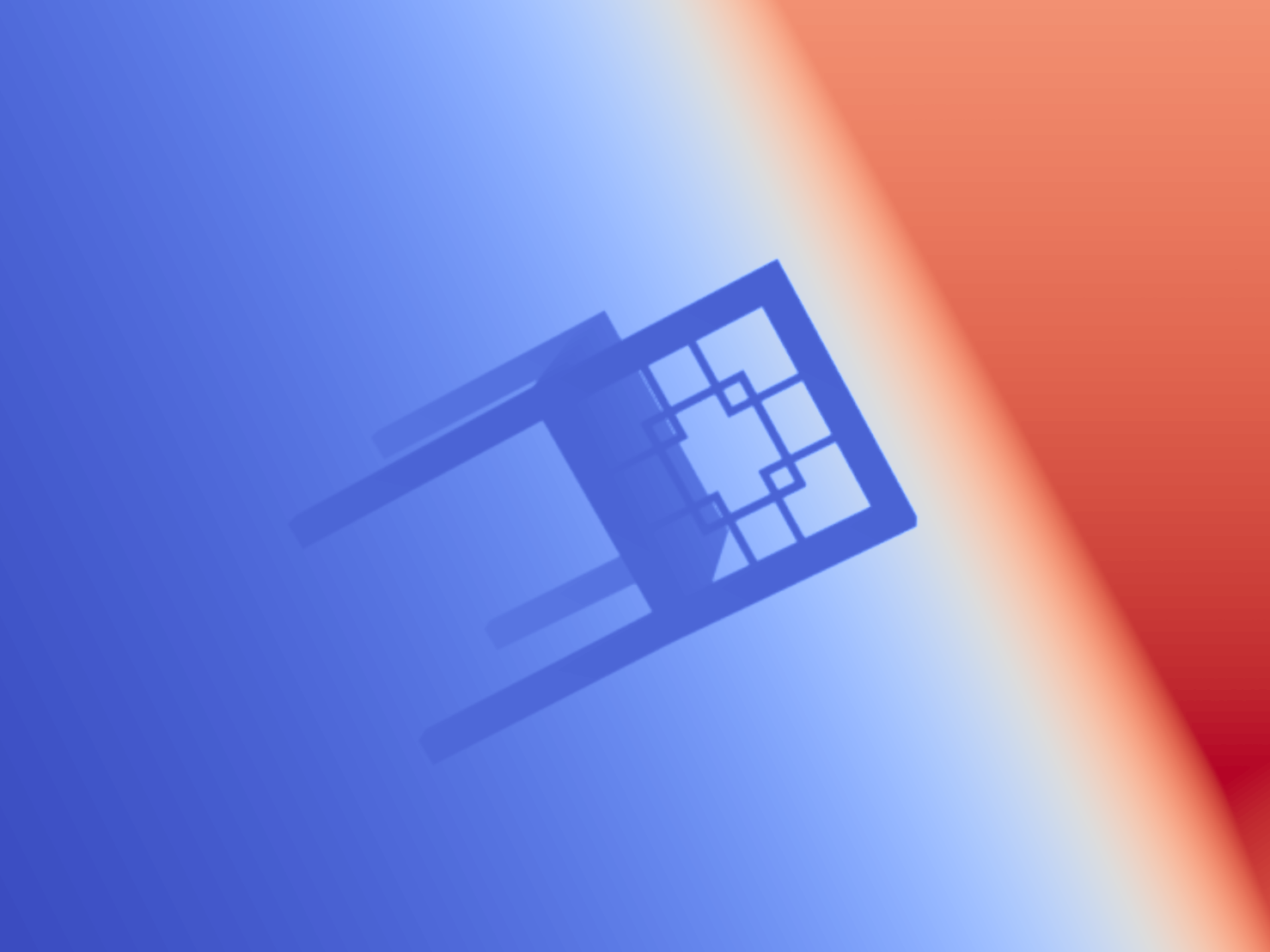}
    \includegraphics[width=3.4cm]{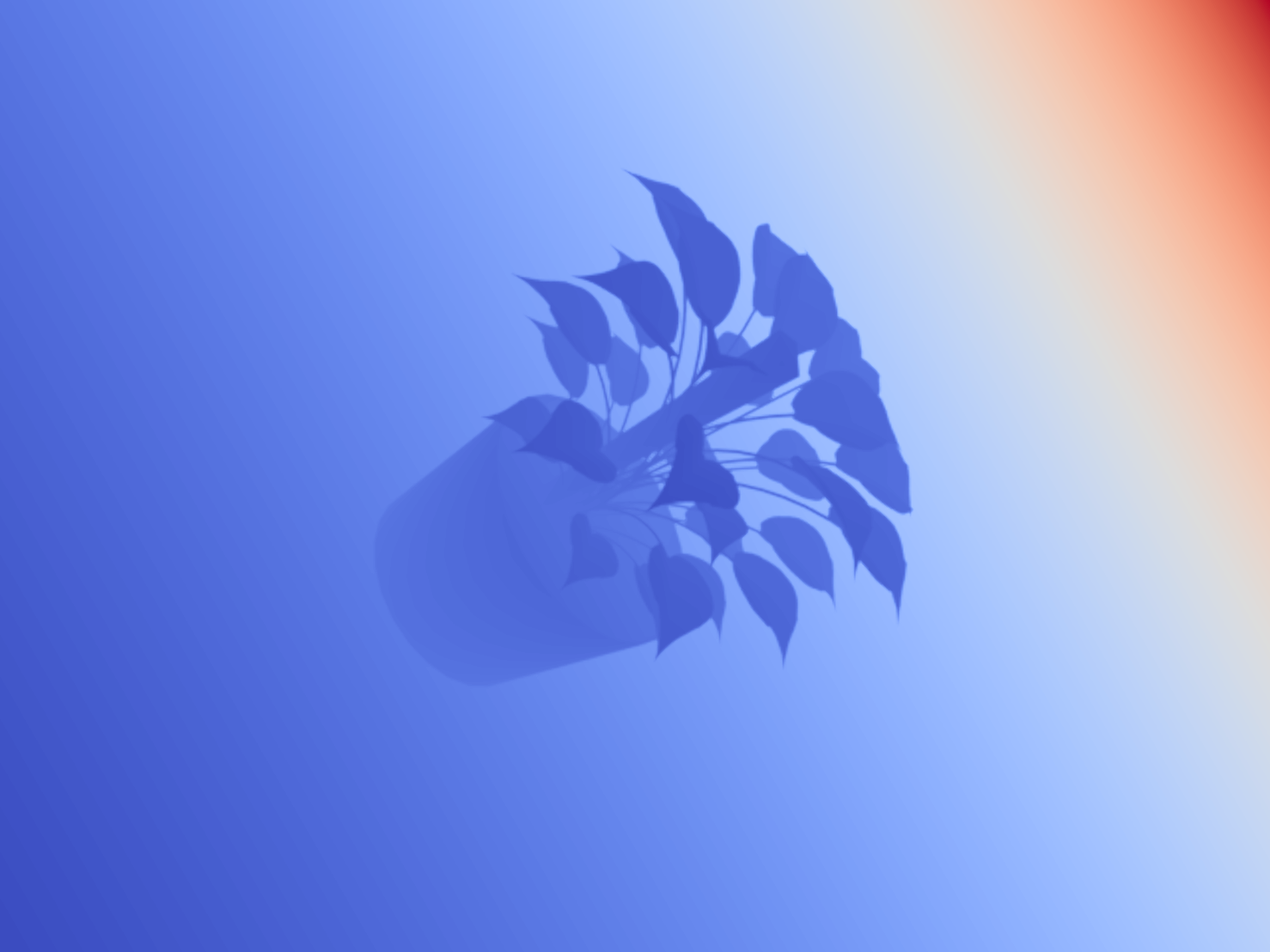}
    \includegraphics[width=3.4cm]{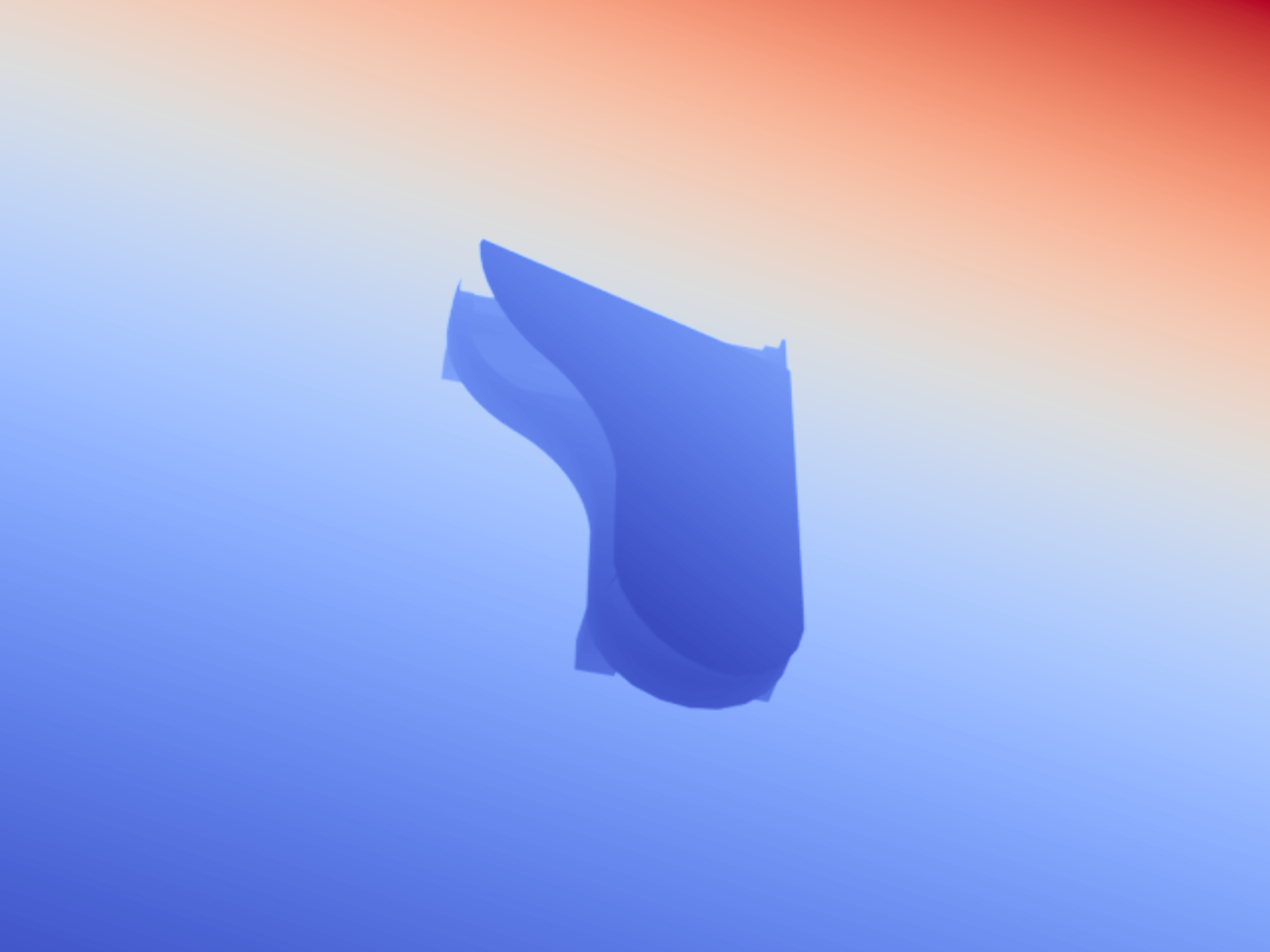}
    \includegraphics[width=3.425cm]{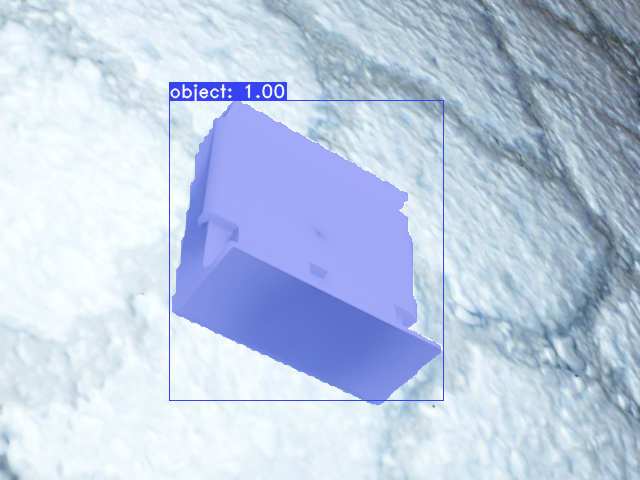}
    \includegraphics[width=3.425cm]{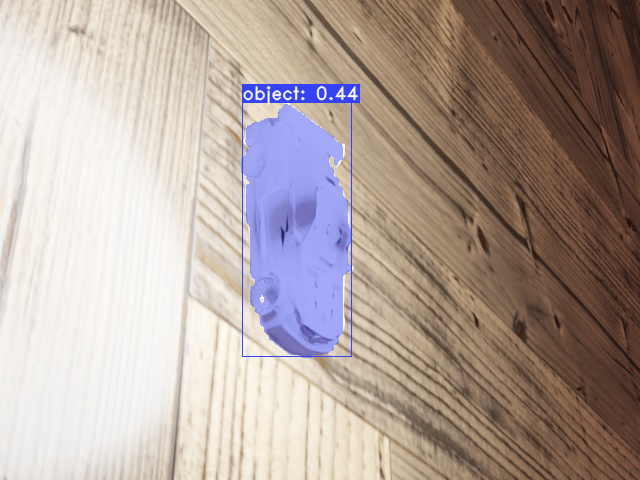}
    \includegraphics[width=3.425cm]{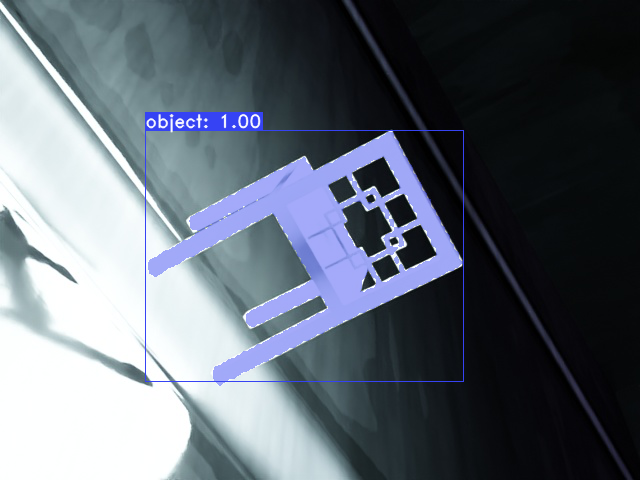}
    \includegraphics[width=3.425cm]{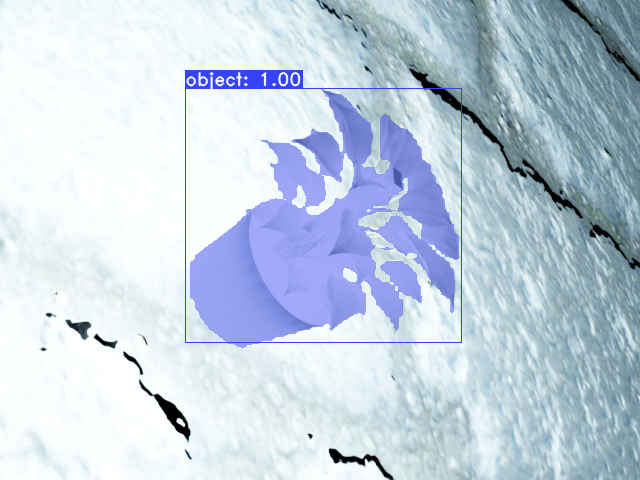}
    \includegraphics[width=3.425cm]{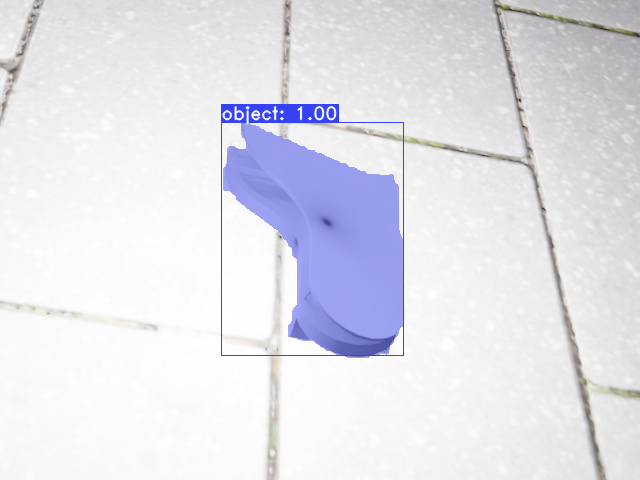}
    \includegraphics[width=3.425cm]{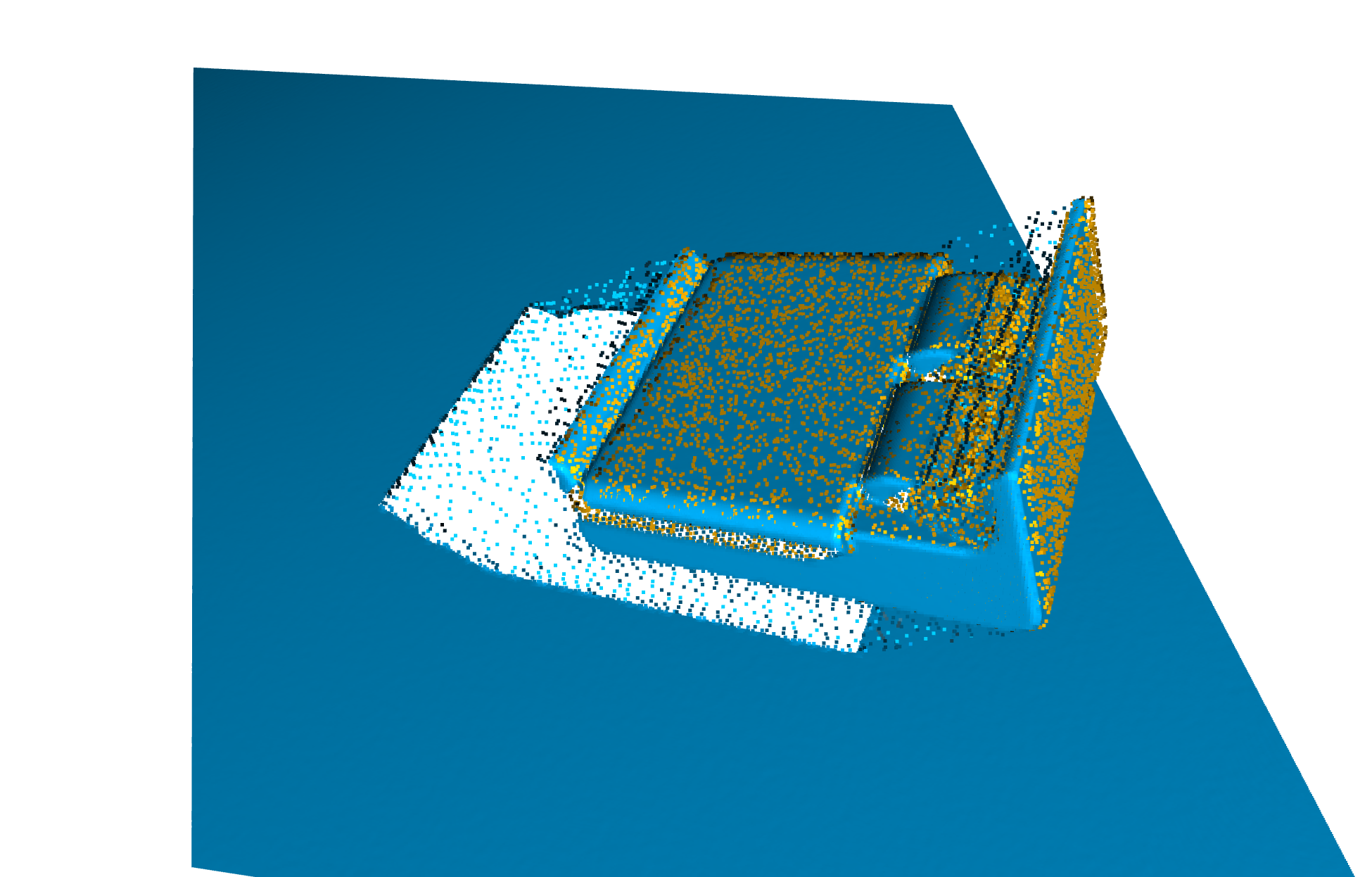}
    \includegraphics[width=3.425cm]{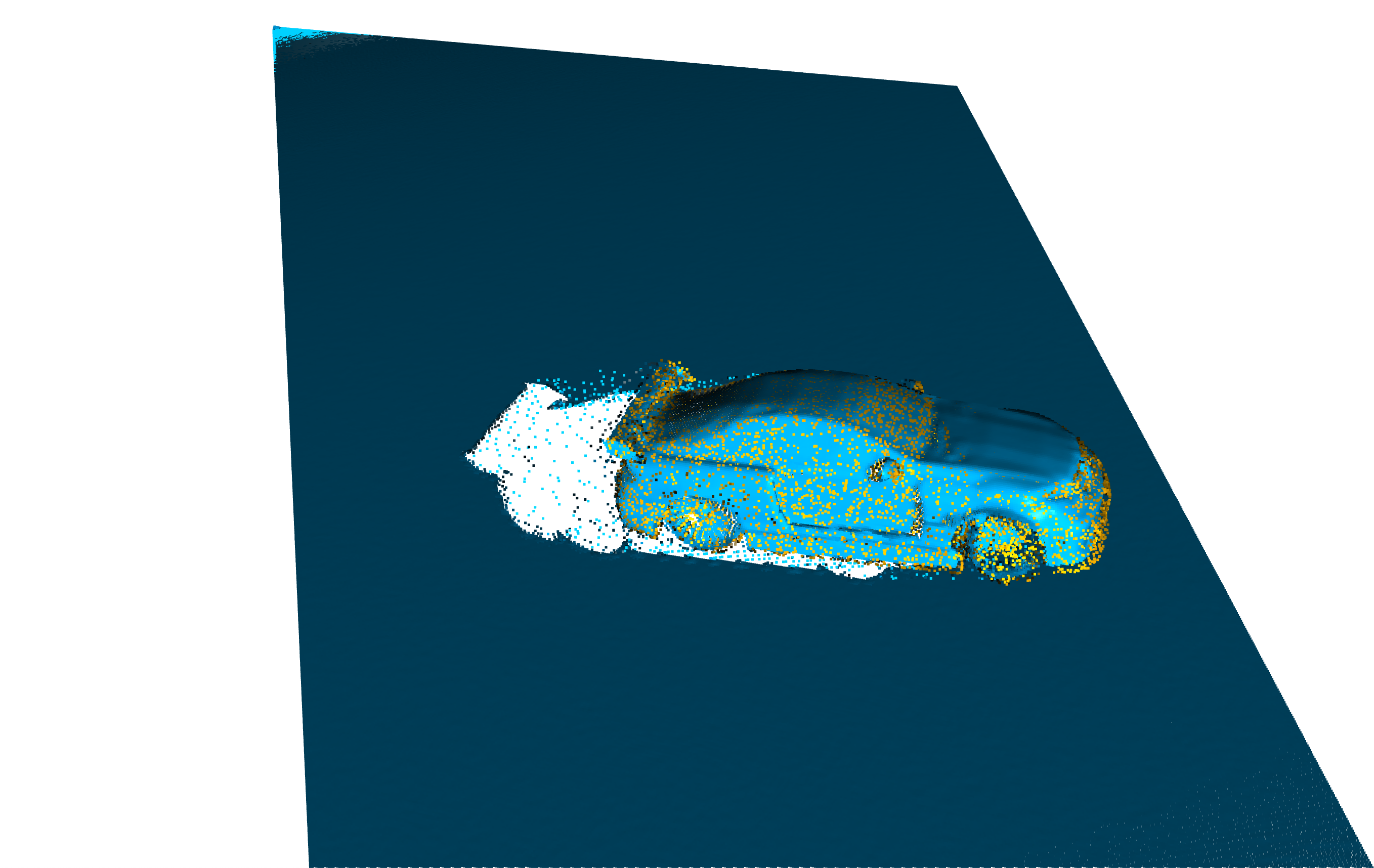}
    \includegraphics[width=3.425cm]{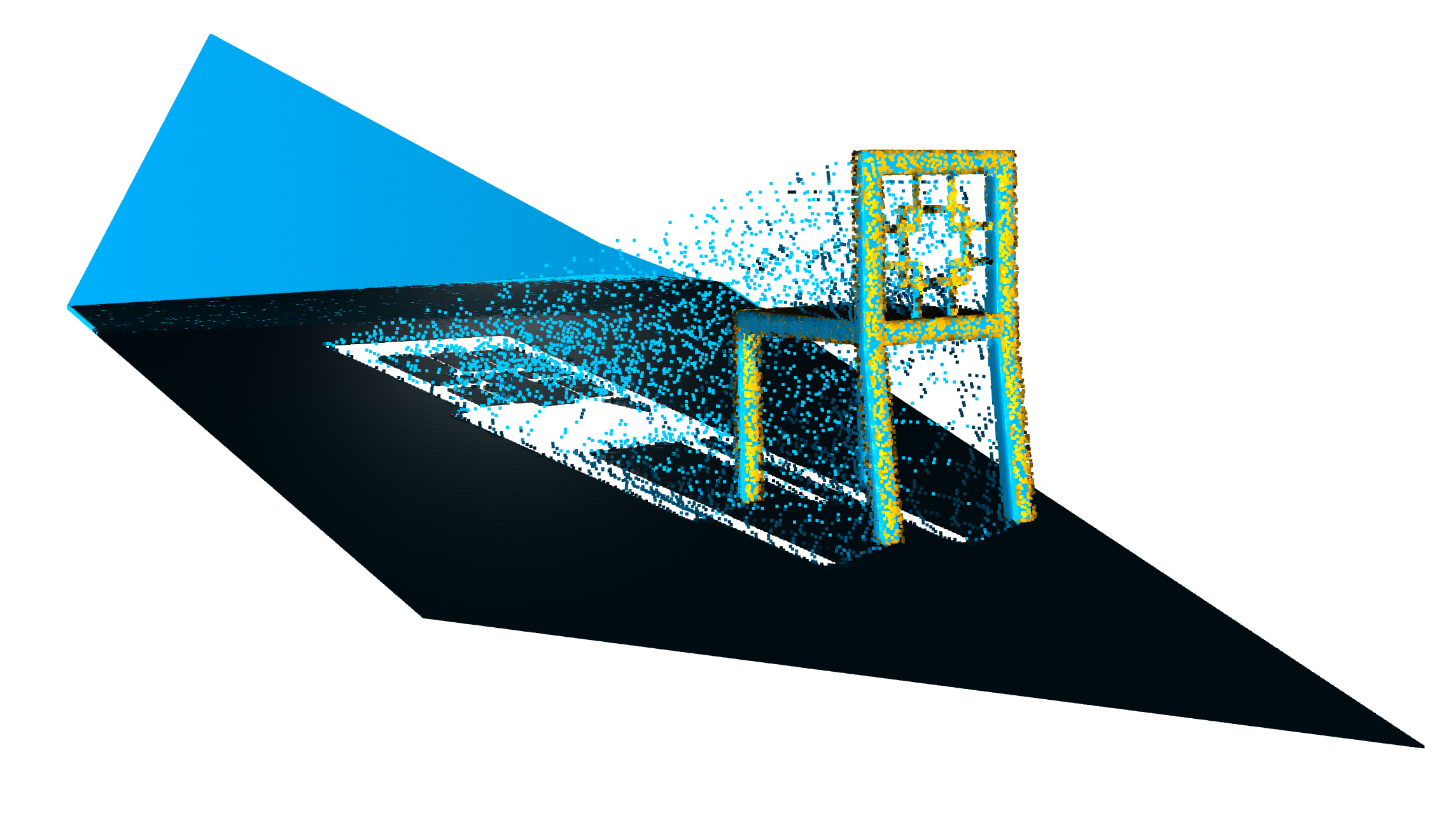}
    \includegraphics[width=3.425cm]{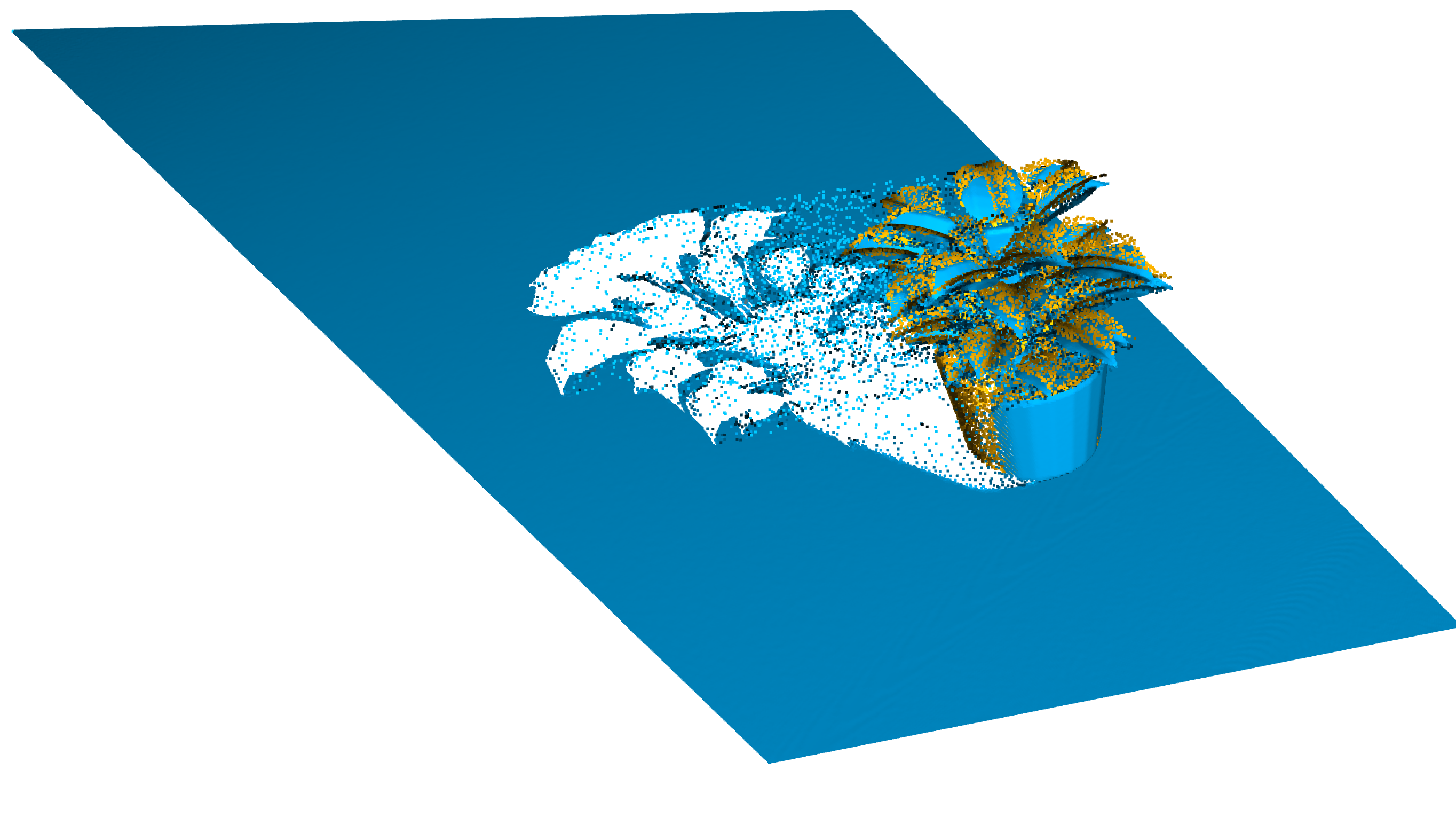}
    \includegraphics[width=3.425cm]{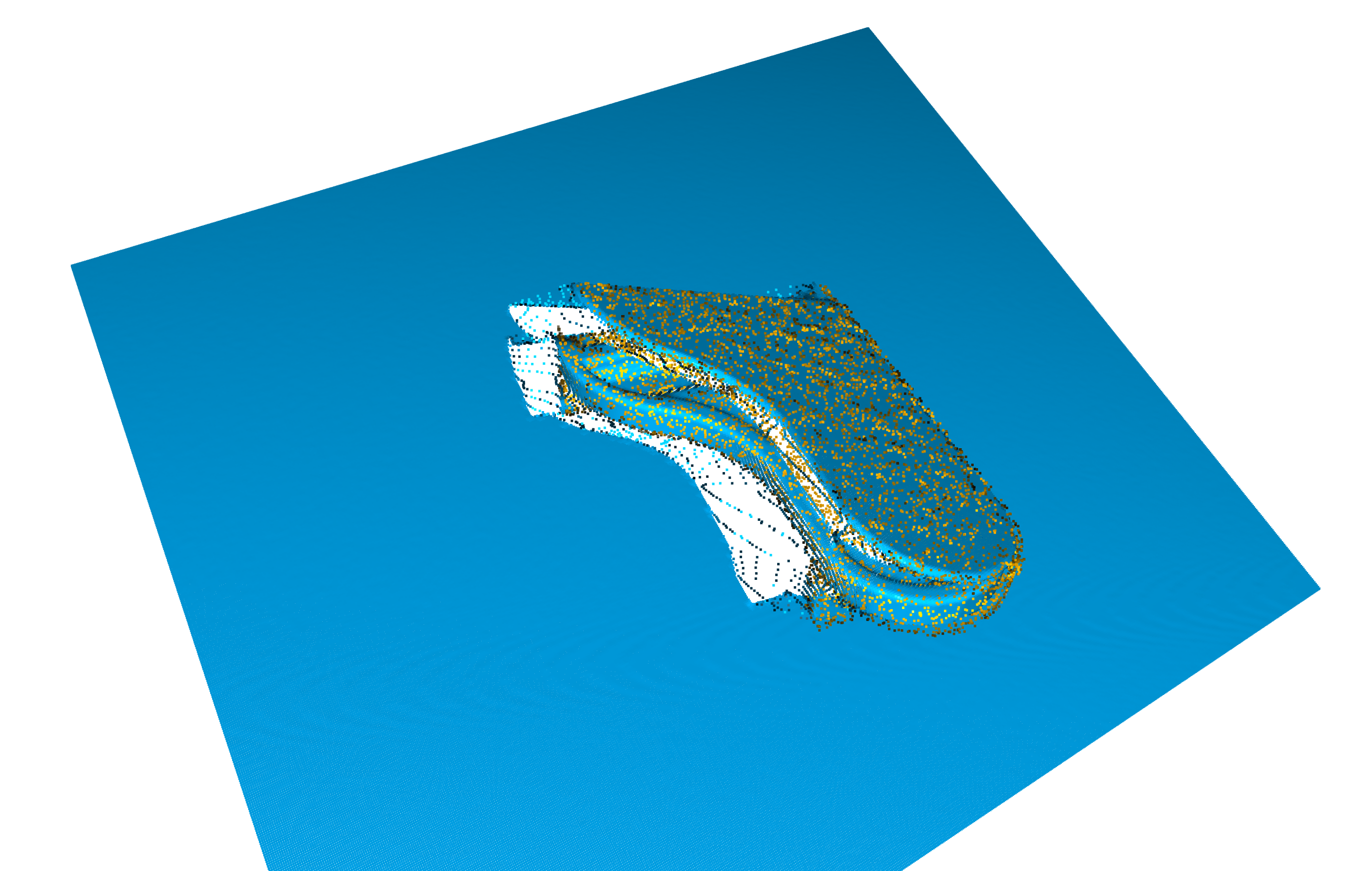}
    
    \caption{\label{synthetic_scene} Qualitative results. (Top) Test depth map used as input to our instance segmentation module. (Middle) Resulting predicted instance mask. (Bottom) Final pose estimation result (Ours-OT).
    }
\end{figure*}

%% file: table/algorithm.tex
\setlength{\textfloatsep}{-1pt}
\begin{algorithm}[!ht]
    \SetAlgoLined
    \SetKwInOut{Input}{Input}
    \SetKwInOut{Output}{Output}
    \SetKwInOut{Init}{Init}
    \SetKwComment{Comment}{$\triangleright$\ }{}
    \Input{
        $\bbS$, $\ba=[\mathbbm{1}_{M}^{\top}, N]^{\top}$, $\bb=[\mathbbm{1}_{N}^{\top}, M]^{\top}$, $\mathbf{f}_{0}=\mathbb{0}_{M+1}$, $\bg_{0}=\mathbb{0}_{N+1}$, \\
        $\lambda$ and $k$ (number of iterations)
        }
    \Output{
        Assignment matrix $\bbP$
        }
    \Init{
        $\bbS = -\bbS / \lambda$
        }
    \While{$l \leq k$}{
        $\mathbf{f}^{(l + 1)} = \lambda log(\ba) - \mathit{logsumexp}(\bbS + \mathbbm{1}_{N+1}\bg^{(l)\top})$ \\%\Comment*[r]{The definition of logsumexp\footnote{$logsumexp(x_{1}, x_{2}, ..., x_{n}) = log(exp(x_{1}) + exp(x_{2}) + ... + exp(x_{n}))$}}
        $\bg^{(l + 1)} = \lambda log(\bb) - \mathit{logsumexp}(\bbS + \mathbf{f}^{(l+1)}\mathbbm{1}_{M+1}^{\top})$ %\Comment*[r]{alternatively update $\mathbf{f}$ and $\bg$.}
    }
    $\bbP_{i,j}=\mathit{exp}(\bbS_{i,j} + \mathbf{f}_i + \mathbf{g}_j)$ \Comment*[r]{transfer back to the original domain.}
    \caption{Log-domain Sinkhorn's algorithm.}
    \label{alg:sinkhorn}
\end{algorithm}

%% file: tex/experiment.tex
\vspace{-0.2cm}
\section{Experiment}
    \input{fig/mAP_noback}
    \input{fig/mAP_withback}
    We first compare our approach to the state-of-the-art methods on the task of estimating the pose of a self-occluded object without a scene background, and then turn to the more challenging scenario where the object is immersed in context. %We test our algorithm both in synthetic scene and real scene. 
    Finally we analyze the influence of the different components of our approach.
    
\subsection{Self-Occluded Object Registration without Background}
    \label{no_background}
    \paragraph{Dataset.} For this experiment, we use the auto-aligned ModelNet40 dataset~\cite{Wu15,Sedaghat16}. This dataset contains mesh models for $40$ object categories. The point clouds are normalized in the range $[-1, 1]$ on each axis. As in~\cite{Wang19f}, we split the data into $9,843$ training and $2,468$ testing mesh models. % \MS{What is a sample here? A specific object?} 
    We use the mesh models to render depth maps given camera viewpoints and intrinsic parameters. Specifically, we treat the full point clouds corresponding to the meshes as source sets $\bX$ and the resulting depth maps as target sets $\bY$. We use the look-at method~\footnote{https://www.scratchapixel.com/lessons/mathematics-physics-for-computer-graphics/lookat-function} to place the camera, and set the distance between the camera and the center of the object to be $0.65$. We then randomly sample the elevation in $[15^{\circ}, 75^{\circ}]$ and azimuth in $[0^{\circ}, 89^{\circ}]$. To deal with the large number of points, following~\cite{Choy20,Yang20}, we first voxelize the point clouds with a voxel size of $0.05 \times (\frac{\sqrt2}{2})$, and then randomly sample the desired number of points from the voxelized point clouds. 
    
 \vspace{-.4cm}
    \paragraph{Evaluation metrics.} We report the rotation error and translation error between the predictions $\hR,\hht$ and the ground truth $\gR,\gt$. These errors are computed as
    \begin{equation}
        \begin{aligned}
            &E_{rot}(\hR,\gR) = arccos\frac{trace(\hR^{\top}\gR) - 1)}{2} \;,\\
            &E_{trans}(\hht,\gt) = \norm{\hht - \gt}^{2}_{2} \;.
        \end{aligned}
    \end{equation}
    We summarize the results in terms of mean average precision (mAP) of the estimated relative pose under varying accuracy thresholds, as in ~\cite{Yi18}. For the rotation, we use the three thresholds $[5^{\circ}, 10^{\circ}, 15^{\circ}]$, and for the translation, we set the thresholds to be $[1\times 10^{-3}, 5\times 10^{-3}, 1\times 10^{-2}]$.
    
    \vspace{-0.4cm}
    \paragraph{Implementation details.}
    We implement our pose estimation network in Pytorch~\cite{Paszke17} and train it from scratch. We use the Adam optimizer~\cite{Kingma15} with a learning rate of $10^{-3}$ and mini-batches of size $20$, and train the network for $40,000$ iterations. We set the number of points for $\bX$ and $\bY$ to be 1024 and 768, respectively, encoding the fact that $\bY$ only contains a visible portion of $\bX$. We use the same parameters for both Ours\_SoftMax and Ours\_OT. For the OT layer, we use $k = 50$ iterations and set $\lambda = 0.5$. Training was performed on one NVIDIA RTX8000 GPU.
    
    To build the ground-truth assignment matrix $\bM$ for Ours\_SoftMax and Ours\_OT, we transform $\bX$ using the ground-truth transformation $\bT$, giving us $\btX$. We then compute the pairwise Euclidean distance matrix between $\btX$ and $\bY$, which we threshold to $0.05$ to obtain a correspondence matrix $\bM \in \{0, 1\}$. For Ours\_OT, we augment $\bM$ with an extra row and column acting as outlier bins to obtain $\bbM$. The points without any correspondence are treated as outliers, and the corresponding positions in $\bbM$ are set to one. This strategy does not guarantee a bipartite matching, which we address using a forward-backward check.
    
    \vspace{-.5cm}
    \paragraph{Registration results.}
    We compare our approach to PRNet~\cite{Wang19f}, DCP~\cite{Wang19e}, ICP~\cite{Besl92a}, FPFH-RANSAC~\cite{Rusu09}, FGR~\cite{Zhou16} and TEASER++~\cite{Yang19,Yang20}. Note that, despite our best effort, we were unable to successfully train DGR~\cite{Choy20} on our dataset. We believe this to be due to the different nature of the task addressed by DGR, i.e., scene to scene registration. This is further evidenced by the fact that, in~\cite{Choy20}, the DGR authors mentioned that they were unable to train DCP on the 3DMatch dataset~\cite{Zeng17b}, whereas we underwent not problems with DCP in our scenario. Similarly, we were unable to train RPMNet~\cite{Yew20} on our data, encountering the instabilities reported by the RPMNet authors, with training crashing after 10 epochs because of an error during SVD.
    
    % About the DGR method, we tried to use the DGR's pre-trained model to fine-tune on our dataset, but we cannot get any reasonable result. 
    % %The main reason is that the object is not like the indoor scene has that rich unique feature. Most of the area on the object is plane area such as the surface of a car, plane or the back of the piano. But indoor scene has a lot of furniture and decorations which is quite unique. 
    % The only result we could get is using the ModelNet40 dataloader which provided in the MinkowskiEngine's example~\cite{Choy19}. That dataloader sample points from each facets on the mesh model instead of uniformly sampled from the mesh model, thus the unique structure of the object is highlighted. But the same strategy cannot be used in our experiment setting. For the reason that the point cloud got from the depth map doesn't have the facet.
    % The DGR paper mentioned that DCP cannot get reasonable result on the 3DMatch dataset~\cite{Zeng17b}. We believe this is the evidence that object level dataset has the different feature distribution with the scene level dataset. We believe that there is still a big gap between global and local feature descriptor, to unified these two descriptor is not a trivial work, but it's out of the scope of this paper.
    
    % About RPMNet method, we tried to retrain the RPMNet on all the categories of ModelNet40. We observed the reported instabilities in the paper, with training crashing after 10 epochs because of an SVD operator error.
    \input{fig/syn_qualitative}
    For the PRNet, we use the pretrained partial-to-partial model provided by the authors, which we fine-tune on our dataset with a learning rate of $0.0001$ for $10$ epochs, where we observed convergence. For DCP, we use the pretrained DCP-v2 model (clean version), and similarly fine-tune it on our dataset with a learning rate of $0.001$ for $30$ epochs corresponding to convergence.
    %\MS{This has nothing to do with the methods, does it? Shouldn't this be in the dataset part?} 
    For ICP, FPFH-RANSAC and FGR, we use the Open3D~\cite{Zhou18} implementations. For TEASER++ we use the official implementation.%~\footnote{https://github.com/MIT-SPARK/TEASER-plusplus}.
    
    The results are summarized in Fig.~\ref{mAP_noback}. Our methods outperforms all the baselines in all settings, with our two variants yielding similar results. 
    TEASER++ achieves the best result among the baselines for a threshold of $5^{\circ}$, but is outperformed by
    DCP for $10^{\circ}$ and $15^{\circ}$, which justifies our choice of DCP as our base network. We believe this to illustrate a downside of exploiting handcrafted features, such as FPFH, as TEASER++ does; while such features are reliable for some objects, they do not generalize well to all of them. By contrast, deep networks can learn to leverage different sources of information for different objects. We attribute the relatively poor performance of PRNet to the fact that it was designed specifically for object-to-object registration.
    
%    yields the worst performance is worse than DCP's performance, probably the PRNet's parameter be tuned too much on the partial-to-partial matching task which at a result lose the generalization ability to another type of task. , but not in all the threshold. Probably for the object with rich features it could handle well, but for the rest objects where the FPFH features fails to generate inlier correspondence, there is nothing this method can do.}
    %\MS{This is very short. Can you make just a couple of comments on these results?}

\subsection{Self-Occluded Object Pose Estimation with Background}
    \label{with_background}
    Let us now turn to the more challenging scenario where the object is immersed in a scene. 
    
    \vspace{-0.5cm}
    \paragraph{Dataset.}
    To this end, we generate a dataset using the BlenderProc~\cite{Denninger19} photo-realistic renderer discussed in Section~\ref{sec:training}. 
    %For each scene we randomly pick one mesh model from ModelNet40 and one texture image from the CC0textures. The texture image is attached to the back-ground plane, to be specific, we form these planes into a cube with length $4$ and put the object inside the cube. For each scene we generate the depth map, mask of the visible part of the object and the pose.
    Specifically, we create cubic scenes of length $4$ with random texture images from the CC0textures dataset~\footnote{https://cc0textures.com}. In each scene, we place an object from the ModelNet40 dataset~\cite{Wu15} with a randomly sampled elevation and azimuth. We then render depth maps from given cameras, together with a mask of the object's visible portion, and the object pose and bounding box. We follow the same procedure to generate the training and test data, using the same object splits and the same sampling range for the look-at camera model as in Section~\ref{no_background}.
    %The test data used in our experiments to evaluate all algorithms was also generated in this manner.
    
    The main difference between this experiment and the previous one is that the algorithms must identify the object in the scene. 
    %This is a quite challenging task for correspondence based method like TEASER++. 
    The whole scene contains $640 \times 480$ points, which after voxelization still leaves between $15,000$ and $300,000$ points. Among these, the object itself covers only $1,000$ to $9,000$ points, depending on the structure of the mesh model.
    %(For example, the keyboard has very thin structure when be seen from a certain view, but the piano has very thick structure.) 
    %The point in the source point set only half of them could have the correspondence for the nature of the task. Even the all the visible part of the source point set could find it's corresponding point in the scene after the voxelization. 
    This corresponds to an inlier rate ranging from $0.5\%$ to $23.5\%$, with $79.41\%$ of the samples having an inlier rate less than $5\%$. 
    %The correspondence based methods have to find the correspondence between these two point sets first and then removing the false matches.
    
    %We aligned the BlenderProc and Pytorch3D, so they will generate the same depth map when giving the same pose. \MS{This seems to be too much detail.}
    
    \vspace{-0.6cm}
    \paragraph{Implementation details.}
    \input{fig/real_scene}
    For our instance segmentation network, we use a pre-trained Resnet101 backbone~\cite{He16}.
    %and use the same hyper-parameter values \MS{what hyper-parameters are we talking about here?} as the base module \MS{What is this?} in~\cite{Bolya19}. \MS{Do you mean that you use the YOLACT model pre-trained on something? If so, just say so. You can still keep the name YOLACT here.}
    Since the input to our model is not an image, we replicate the depth map three times to match the number of input channels. As output, we keep only one category, which is used to classify each pixel as belonging to the object or not. We train the resulting network for $100,000$ iterations with mini-batches of size $16$, and using SGD with a learning rate of $10^{-3}$.
    We then train our pose estimation module in the same manner as in Section~\ref{no_background}, but discarding the points that do not belong to the object using the ground-truth mask. This, of course, can only be achieved during training, and leads to a mismatch between training and testing time, since, at testing time, we cannot expect our instance segmentation module to always give perfect results. To handle this, at test time, we erode the boundary of the predicted masks with a $3\times 3$ kernel. Furthermore, we remove all the points that lie outside a sphere centered on the object and of radius $1.2$, accounting for the fact that the object coordinates were normalized in the range $[-1, 1]$.

    \vspace{-0.4cm}
    \paragraph{Registration results.}
    We compare our approach to the same baselines as in Section~\ref{no_background}, using the same hyper-parameter values to train them. The only exception to this is for TEASER++, for which we limit the number of points in the target point set to a maximum of $50,000$ because we observed too large target point sets 
    %larger than $100,000$ the algorithm will 
    to produce a segmentation fault in the code.
    %Ours\_OT and Ours\_SoftMax methods are using the instance segmentation module's output mask. 
    Note that, prior to our work, no method was designed to explicitly estimate the 6D pose of a self-occulded object observed in a scene. Therefore, the baselines directly tackle the 3D registration task from the raw depth maps. By contrast, our segmentation module outputs a mask that facilitates the task of the registration one.
    %For the rest of the method, they take the raw depth map and transfer to the target pointcloud then using voxelization method to downsample the pointcloud to it required number. 
    
    The result are summarized in Fig.~\ref{mAP_withback}, with a few examples in Fig.~\ref{fig:syn_qualitative}. Note that our methods outperform the baselines by a large margin. Since the DCP and PRNet directly act on the raw depth maps, the point-cloud downsampling process, required to make inference tractable, leaves only very few points belonging to the object, thus leading to the failure of these methods. By contrast, TEASER++, which was designed to handle such large amounts of outliers, yields reasonable results, nevertheless clearly outperformed by our learning-based strategy.
%    our main baseline method could yield some reasonable results in this experiment setting, which is an evidence that our task is similar to the full object matching task in TEASER++ but more challenging.}
    % \MS{Again, add a few comments about these results.}

\input{fig/mAP_real}
\subsection{Dealing with Real Data}
    To evidence that our approach generalizes to real depth maps, we use the TUD-L dataset~\cite{Hodan18}, which contains training and testing image sequences that show three moving objects under eight lighting conditions. We retrain our complete model with OT layer from scratch on this dataset, keeping the same hyper-parameter values as for the previous experiments.
    To compare our method to the baselines, we select 2000 images from each testing sequence. We restrict our baseline evaluation to the methods that gave reasonable results in the experiment of Section~\ref{no_background}, i.e., TEASER++ and FGR. As can be shown in Fig.~\ref{mAP_real}, our method also outperforms the baselines by a large margin in this scenario. 
    %all other baselines. In this task, TEASER++ could only have very small amount of samples pass the low error threshold. FGR totally failed in this experiment.}
    In Fig.~\ref{real_scene}, we show a few qualitative results obtained with our framework, evidencing the accuracy of our predictions. 
    %\MS{Why don't we report quantitative results on this dataset? And why don't we compare our results to other methods? Considering that this is part of the BOP challenge, this will seem strange.}

\input{table/inference_time}
\subsection{Method Analysis}
    \paragraph{Inference Time.}
    In Table~\ref{tab:inference_time}, we compare the inference time of our registration method to that of the baselines for different point set sizes. For this comparison to be fair, we ran all methods on a desktop computer with an Intel(R) Core(TM) i7-7700K CPU @ 4.20GHz, an Nvidia GTX 2080 Ti GPU, and 32GB memory. 
    %For the non learning-based methods (ICP and FGR), we used their Open3D~\cite{Zhou18} implementation. For TEASER++, we used the code released by the authors. \MS{Put the link as a footnote.} For the learning-based ones (DCP and PRNet), we used Pytorch implementations. 
    Theoretically, the inference time for Ours\_SoftMax time is the same as DCP, thus we omit it. Ours\_OT yields run-times comparable to FGR, while being slightly faster than PRNet. Note that the run-times reported in Table~\ref{tab:inference_time} for our approach do not include the instance segmentation module, which takes 0.0285s to process an image of size $480\times 640$. 
    %\MS{You should add this value to our run-times in the table, and thus report Ours\_SoftMax as well.}
    
    % \textbf{Effectiveness of rendering on-the-fly.}
    % Here we replace the rendering on-the-fly with the partial-to-partial matching dataset which is used in PRNet. We summarize the result in the Fig.~\ref{mAP_withback} labeled as Ours\_OT\_Partial and Ours\_SoftMax\_Partial. Even our models which trained on this dataset could not give any reasonable result.
    
    \vspace{-.3cm}
    \paragraph{Effectiveness of the Optimal Transport Layer.} Note that the best results with our approach were obtained without the optimal transport layer. Here, we illustrate that this layer is nonetheless important in scenarios where the training data is noisy. To this end, we replace our rendering on-the-fly procedure with BlenderProc during training so as to generate noisy training observations. To be specific, we use the ground-truth mask to obtain the depth of the object and transform it to the target point cloud, and then jitter the rotation in the range $[0, 45^{\circ}]$ and the translation in the range $[-1, 1]$ on each axis. Only Our\_OT could be trained on the resulting dataset, thanks to its outlier bins. We provide the corresponding results as Ours\_OT\_BlenderProc in Fig.~\ref{mAP_withback}.

%% file: fig/mAP_noback.tex
\begin{figure*}[t]
    \centering
    \setlength{\abovecaptionskip}{-.1pt}
    \setlength{\belowcaptionskip}{-10pt}
    \includegraphics[width=8.6cm]{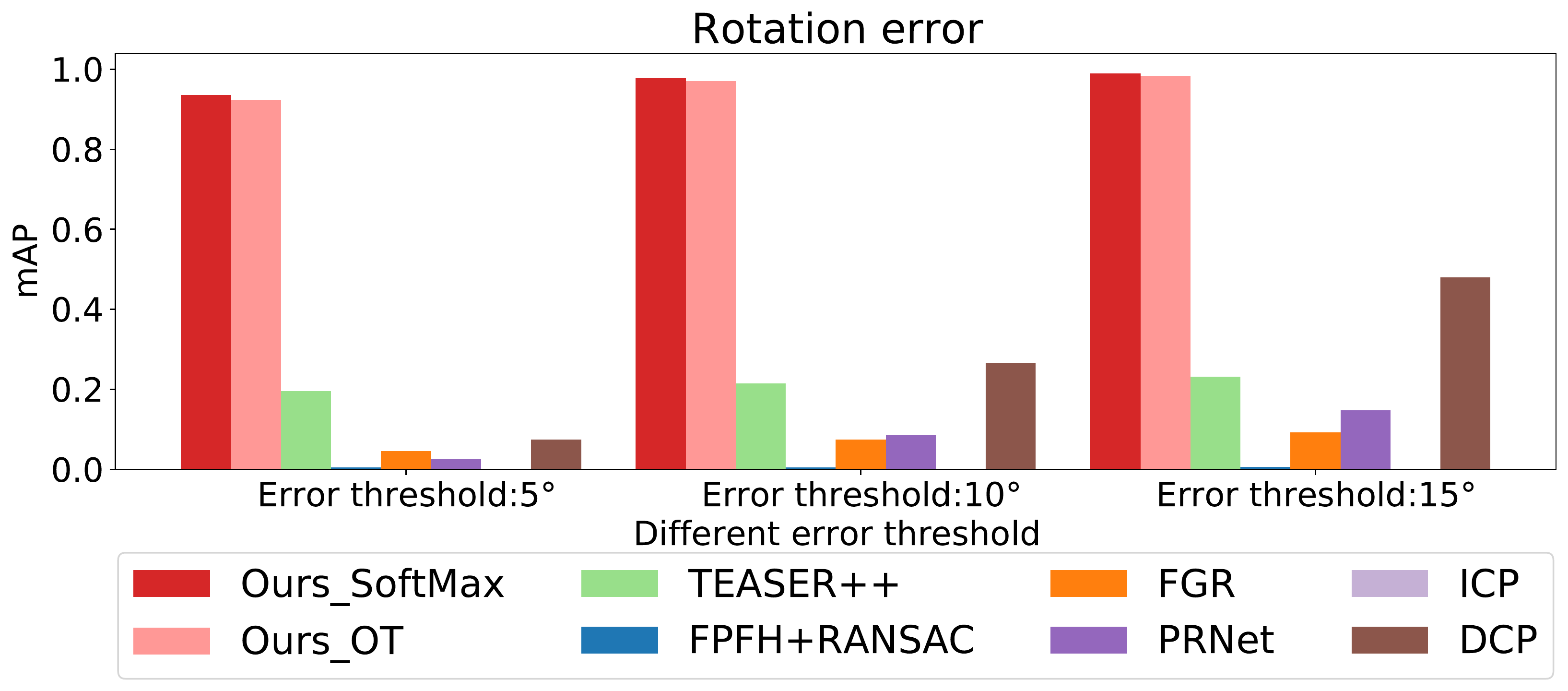}
    \hspace{.1cm}
    \includegraphics[width=8.6cm]{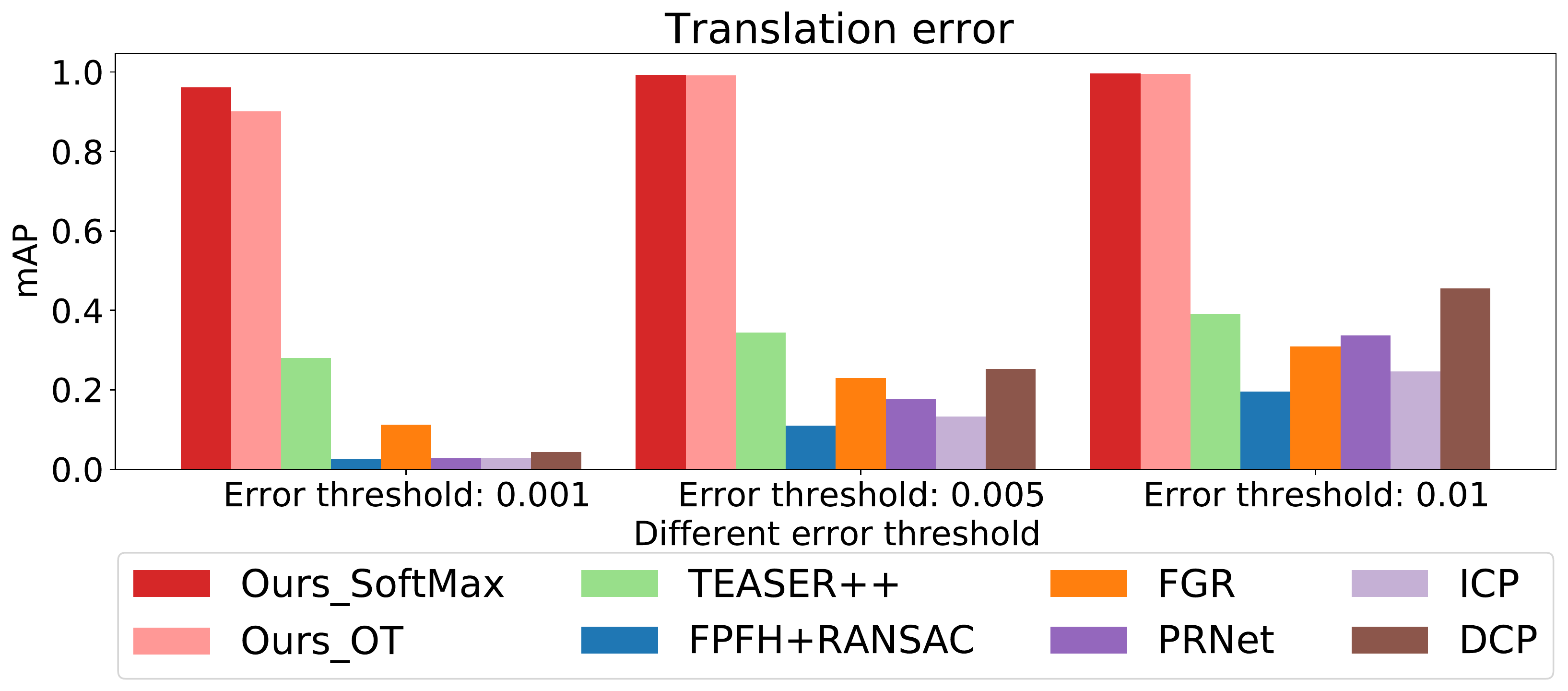}
    \caption{\label{mAP_noback} Rotation and translation mAP for different thresholds on the ModelNet40 dataset without background.}
\end{figure*}

%% file: fig/mAP_withback.tex
\begin{figure*}[t]
    \centering
    \setlength{\abovecaptionskip}{-.1pt}
    \setlength{\belowcaptionskip}{-10pt}
    \includegraphics[width=8.6cm]{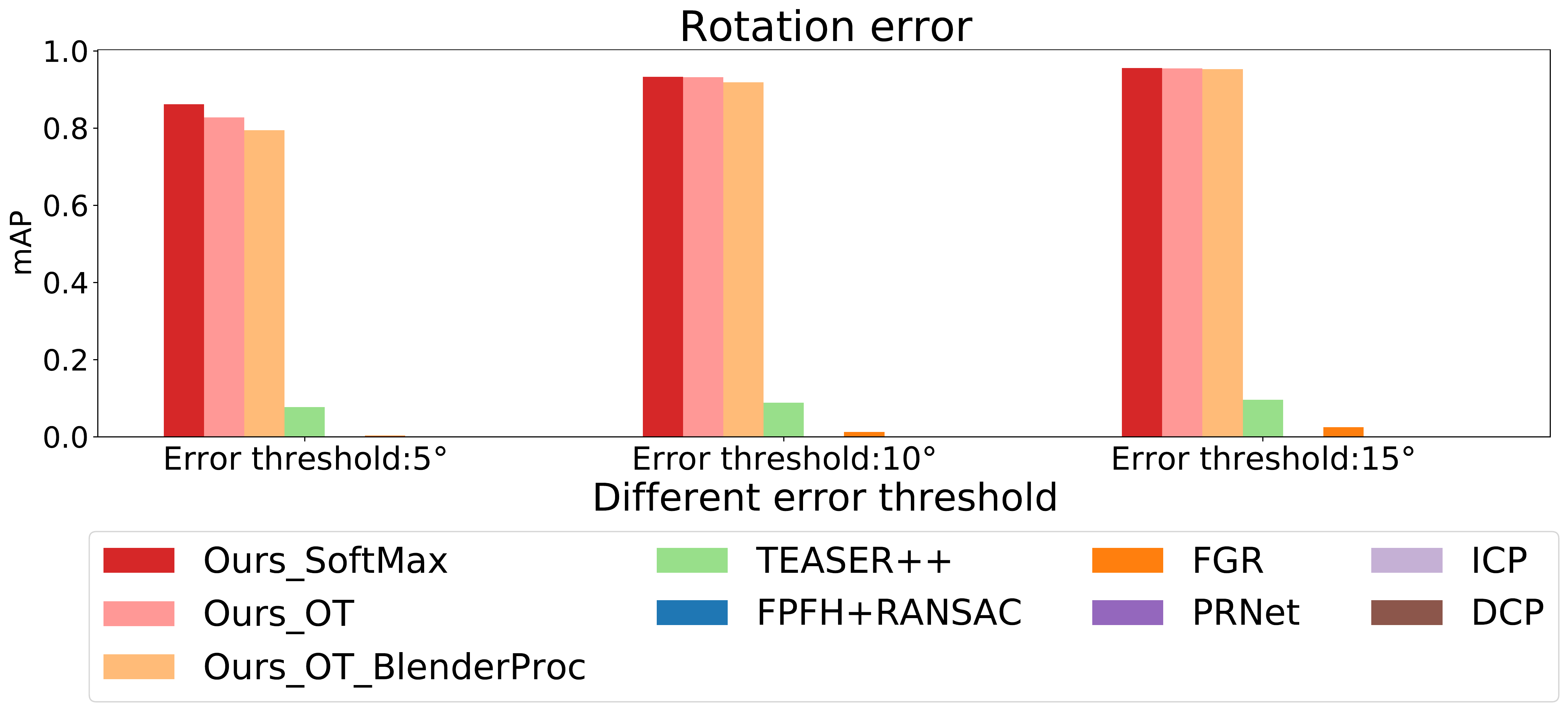}
    \hspace{.1cm}
    \includegraphics[width=8.6cm]{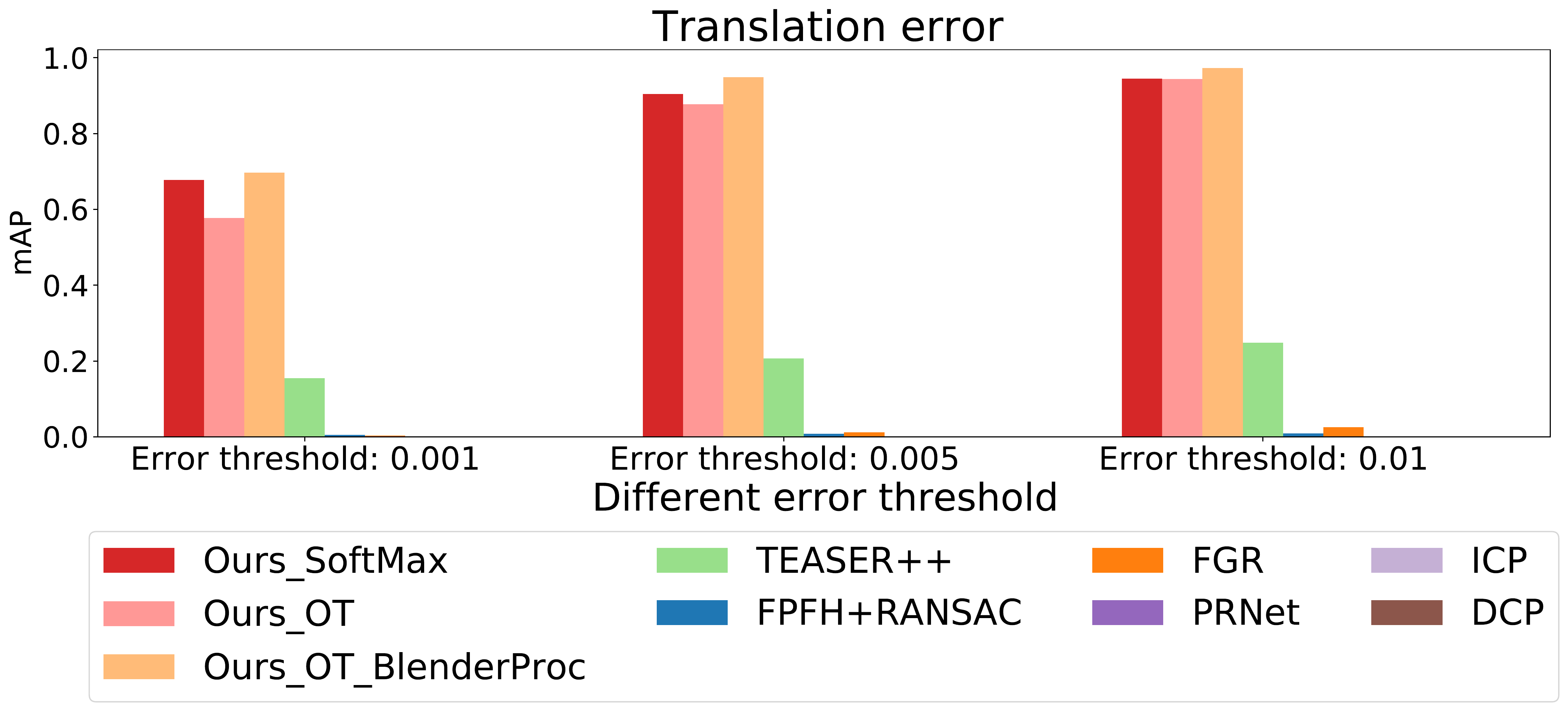}
    \caption{\label{mAP_withback} Rotation and translation mAP for different thresholds on the ModelNet40 dataset with background.}
\end{figure*}

%% file: fig/syn_qualitative.tex
\begin{figure}[t]
    \centering
    \begin{subfigure}[b]{0.47\textwidth}
    \centering
    \includegraphics[width=4cm]{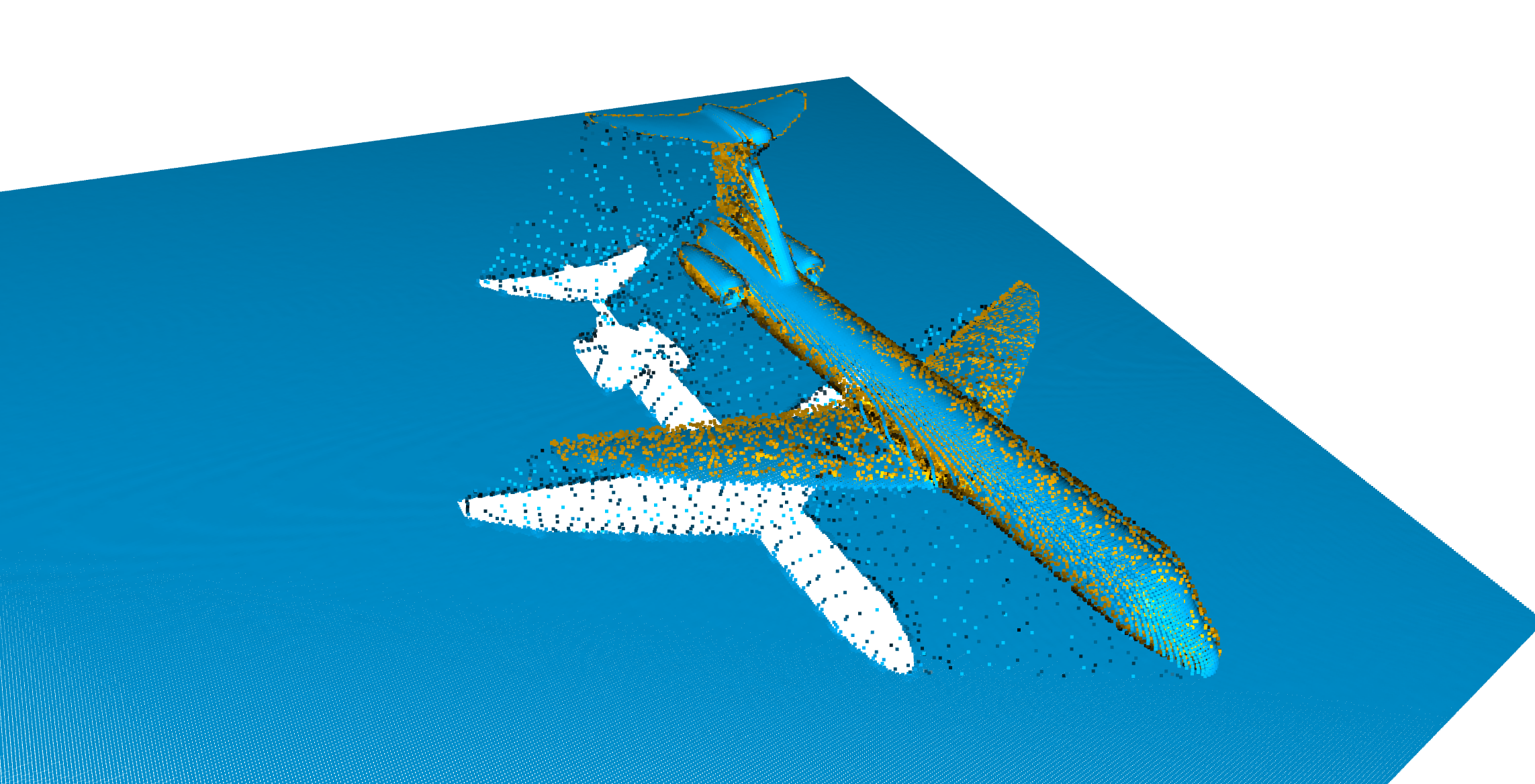}
    \includegraphics[width=4cm]{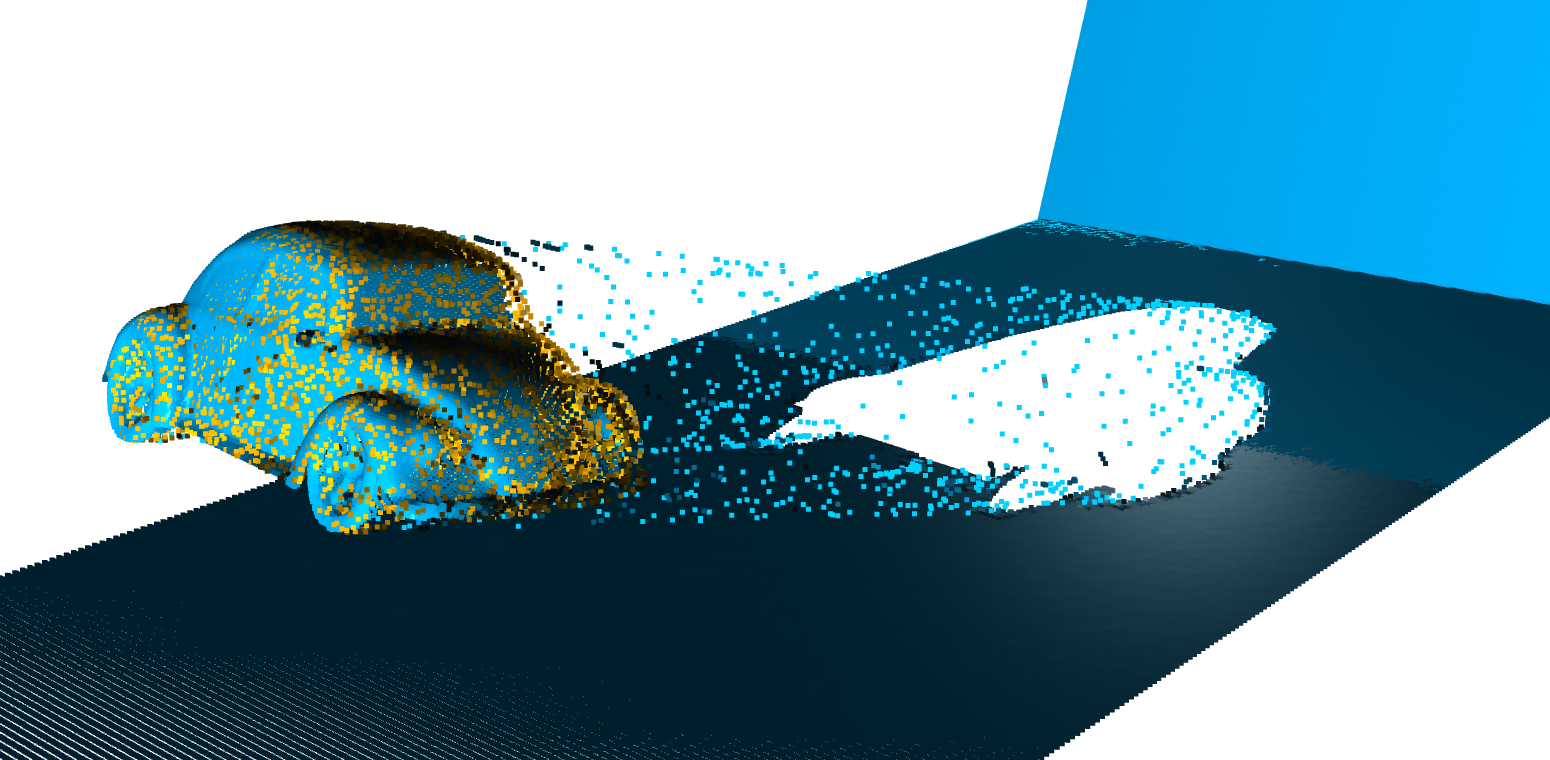}
    \caption{Ours-OT}
    \end{subfigure}
    
    \begin{subfigure}[b]{0.47\textwidth}
    \centering
    \includegraphics[width=4cm]{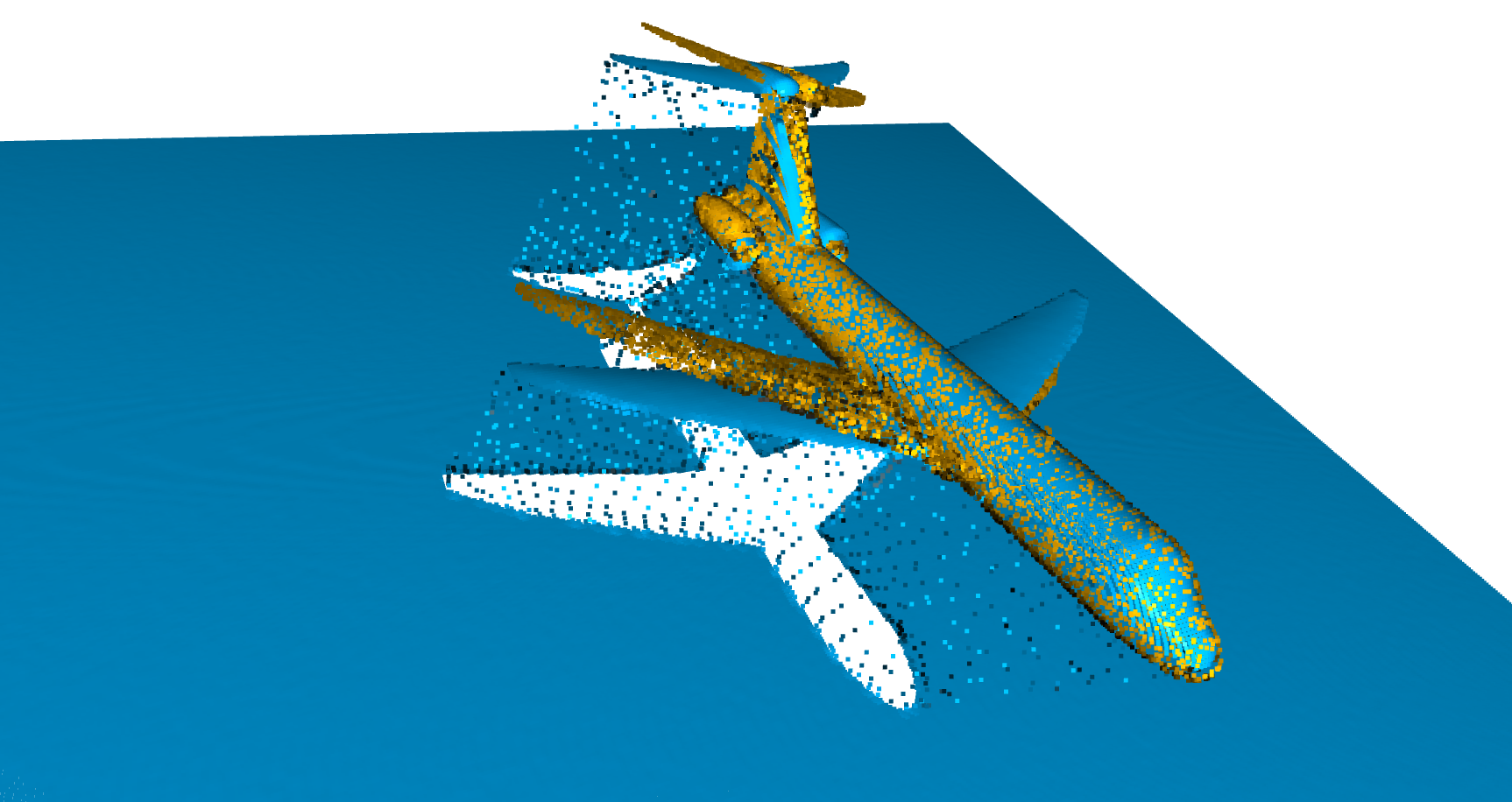}
    \includegraphics[width=4cm]{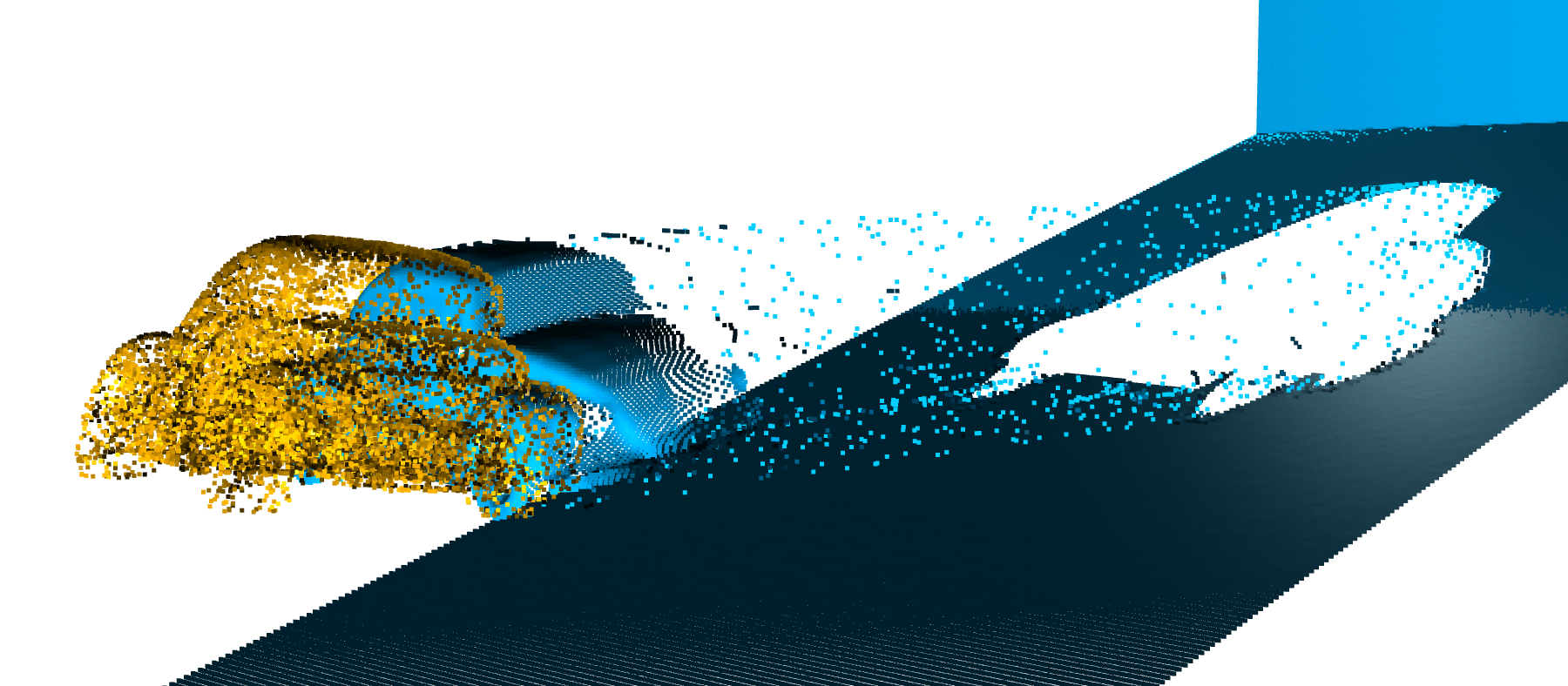}
    \caption{TEASER++}
    \end{subfigure}
    
    \begin{subfigure}[b]{0.47\textwidth}
    \centering
    \includegraphics[width=4.1cm]{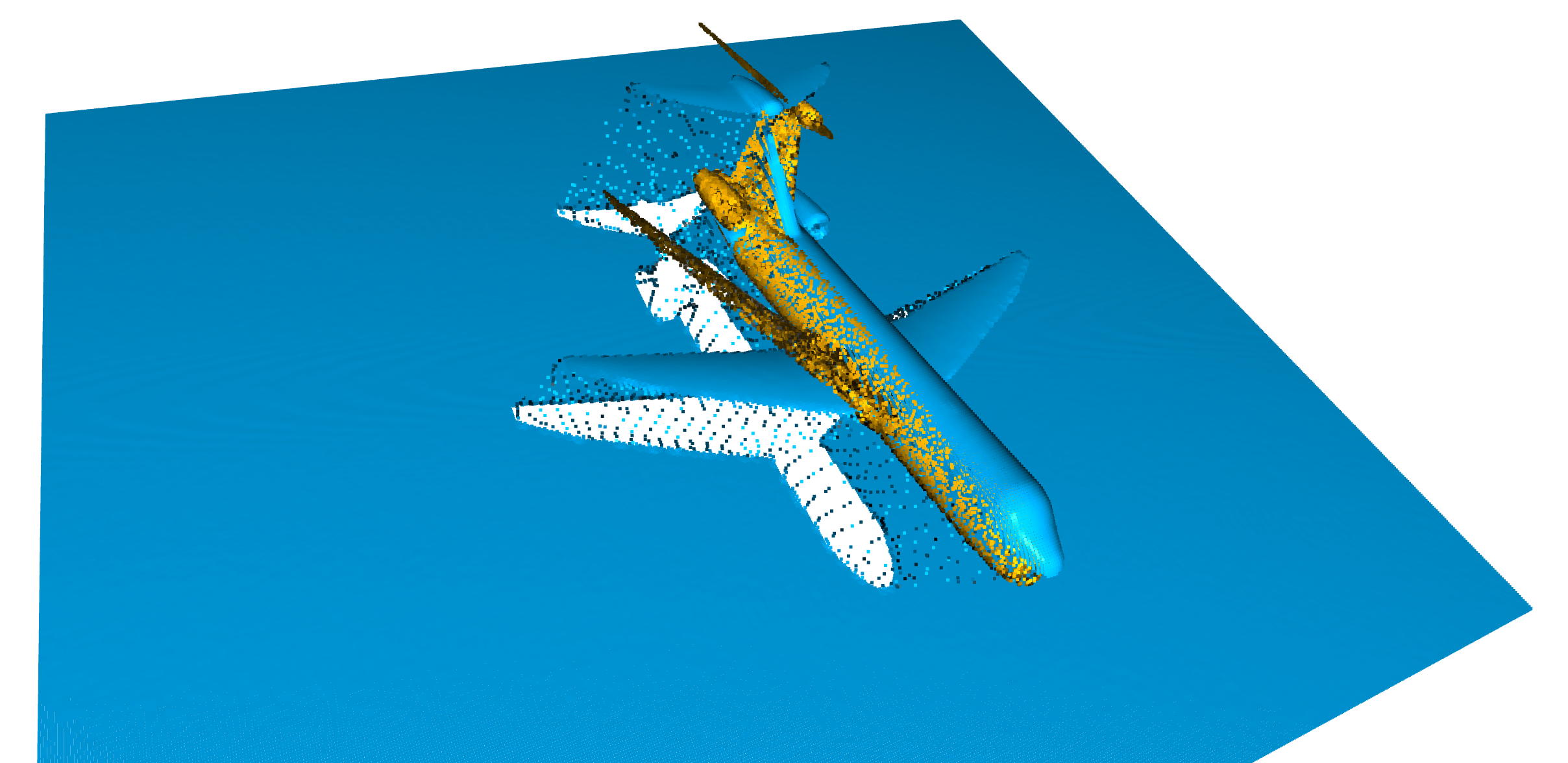}
    \includegraphics[width=4cm]{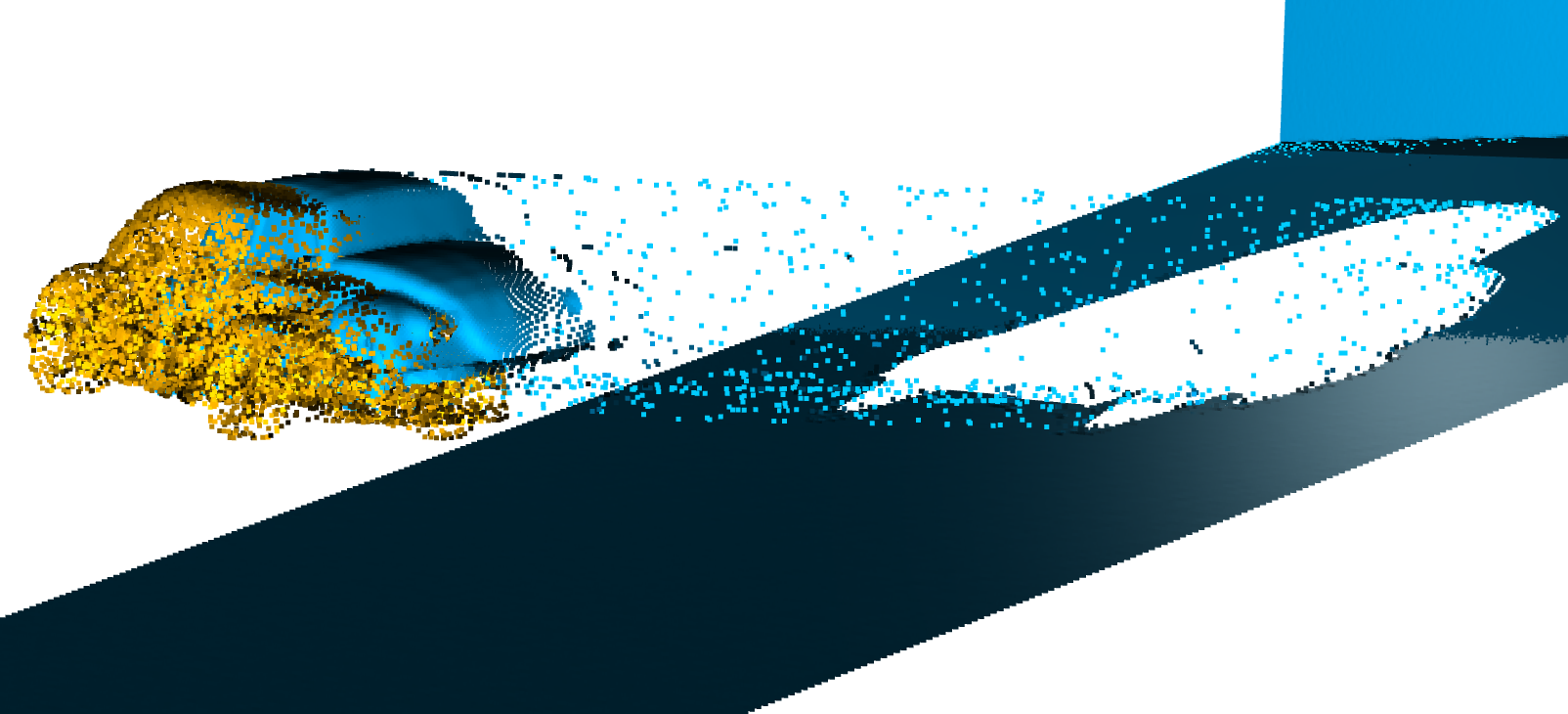}
    \caption{FGR}
    \end{subfigure}
    \setlength{\abovecaptionskip}{-7pt}
    \setlength{\belowcaptionskip}{5pt}
    \caption{\label{fig:syn_qualitative} Qualitative results on the BlenderProc dataset. Note that our approach (Ours-OT) yields much more precise registrations than the two top-performing baselines.}
\end{figure}

%% file: fig/real_scene.tex
\begin{figure}[!ht]
    \centering
    \setlength{\belowcaptionskip}{-8pt}
    \includegraphics[width=3.8cm]{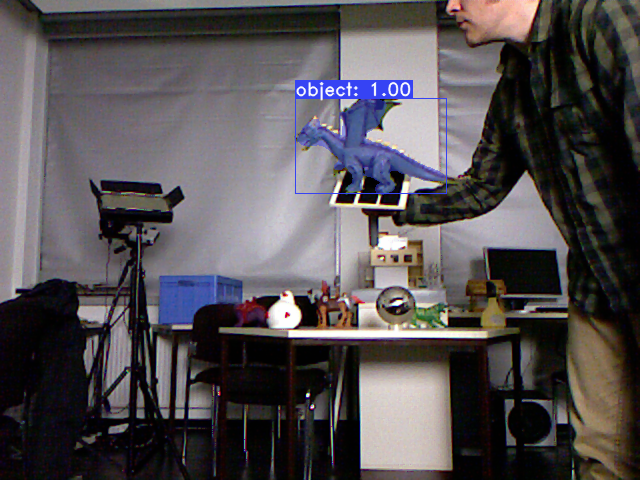}
    \hspace{.2cm}
    \includegraphics[width=4.cm]{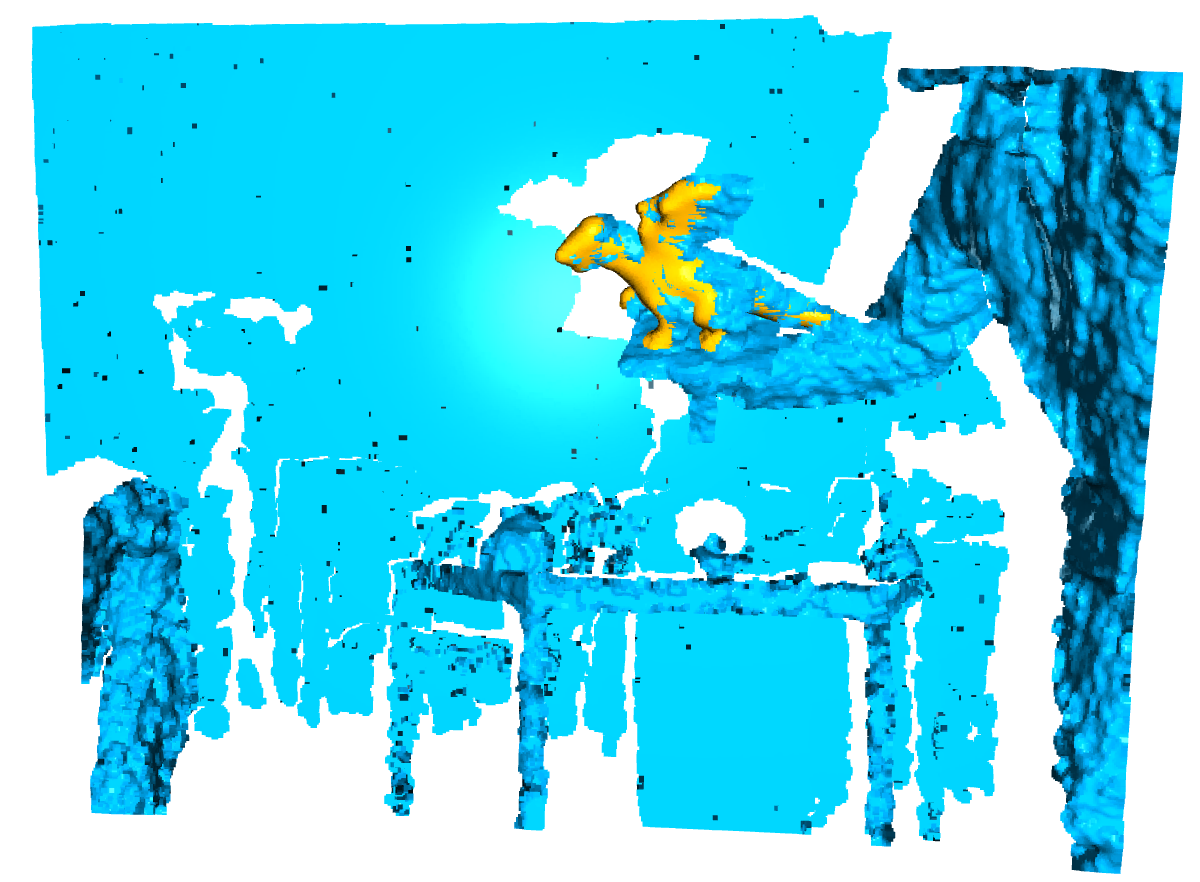}
    \includegraphics[width=3.8cm]{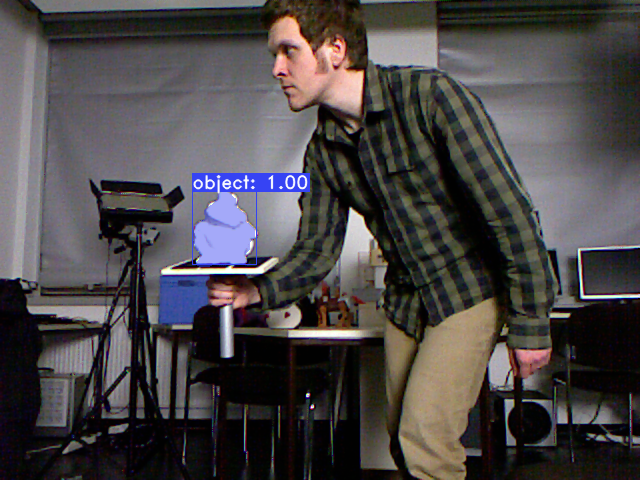}
    \hspace{.2cm}
    \includegraphics[width=4.cm]{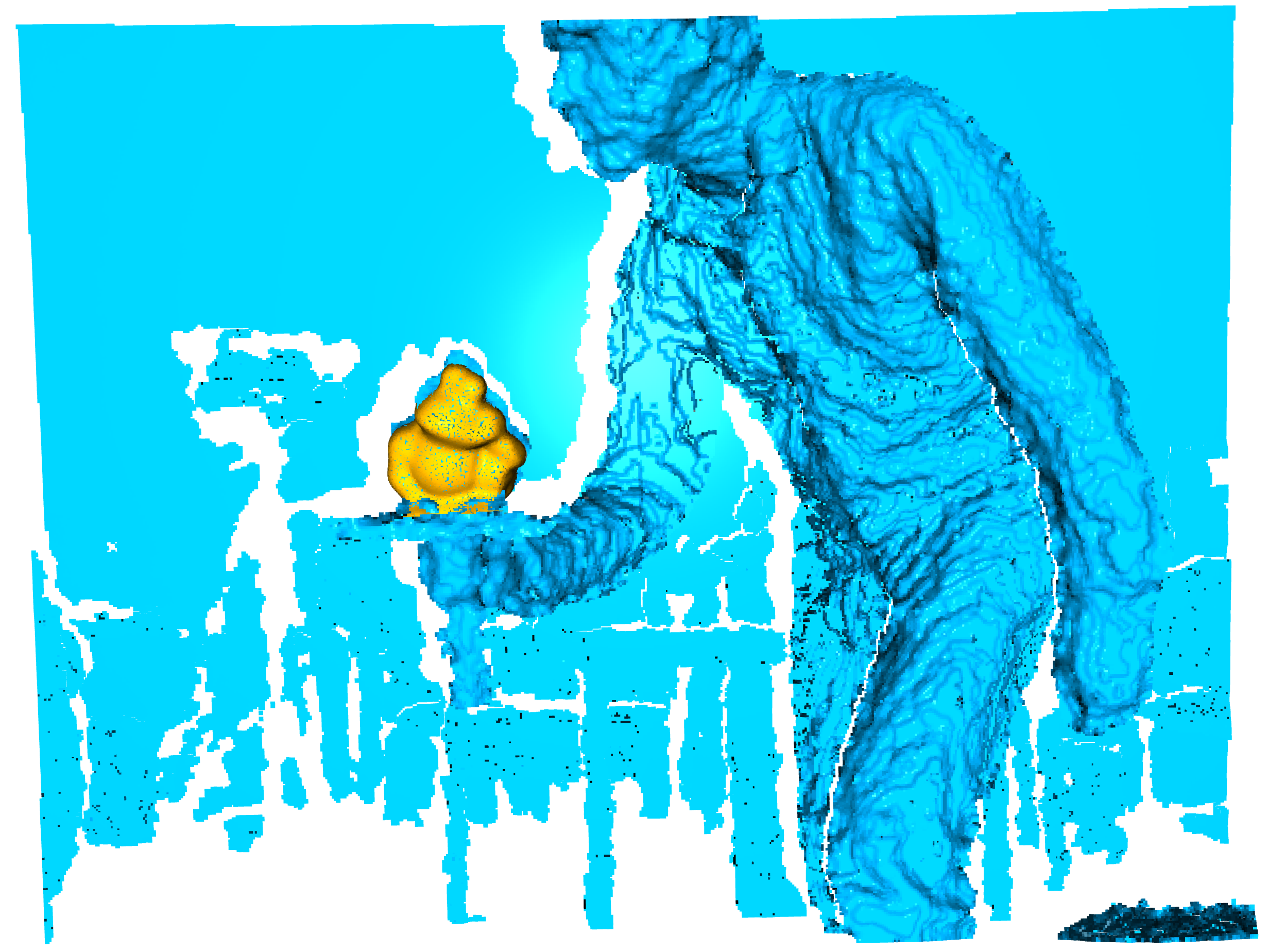}
    \includegraphics[width=3.8cm]{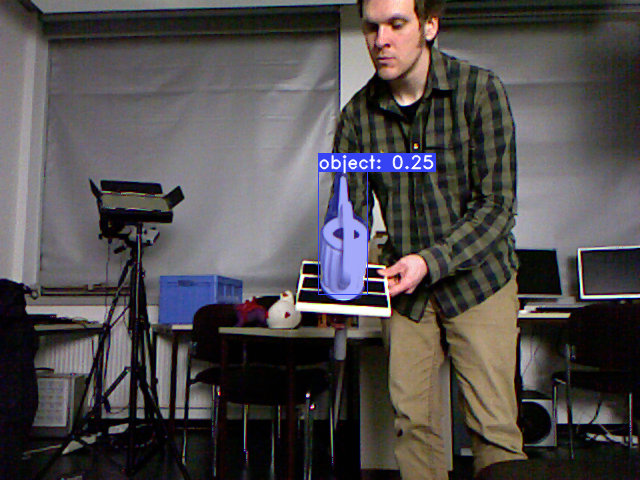}
    \hspace{.2cm}
    \includegraphics[width=4.cm]{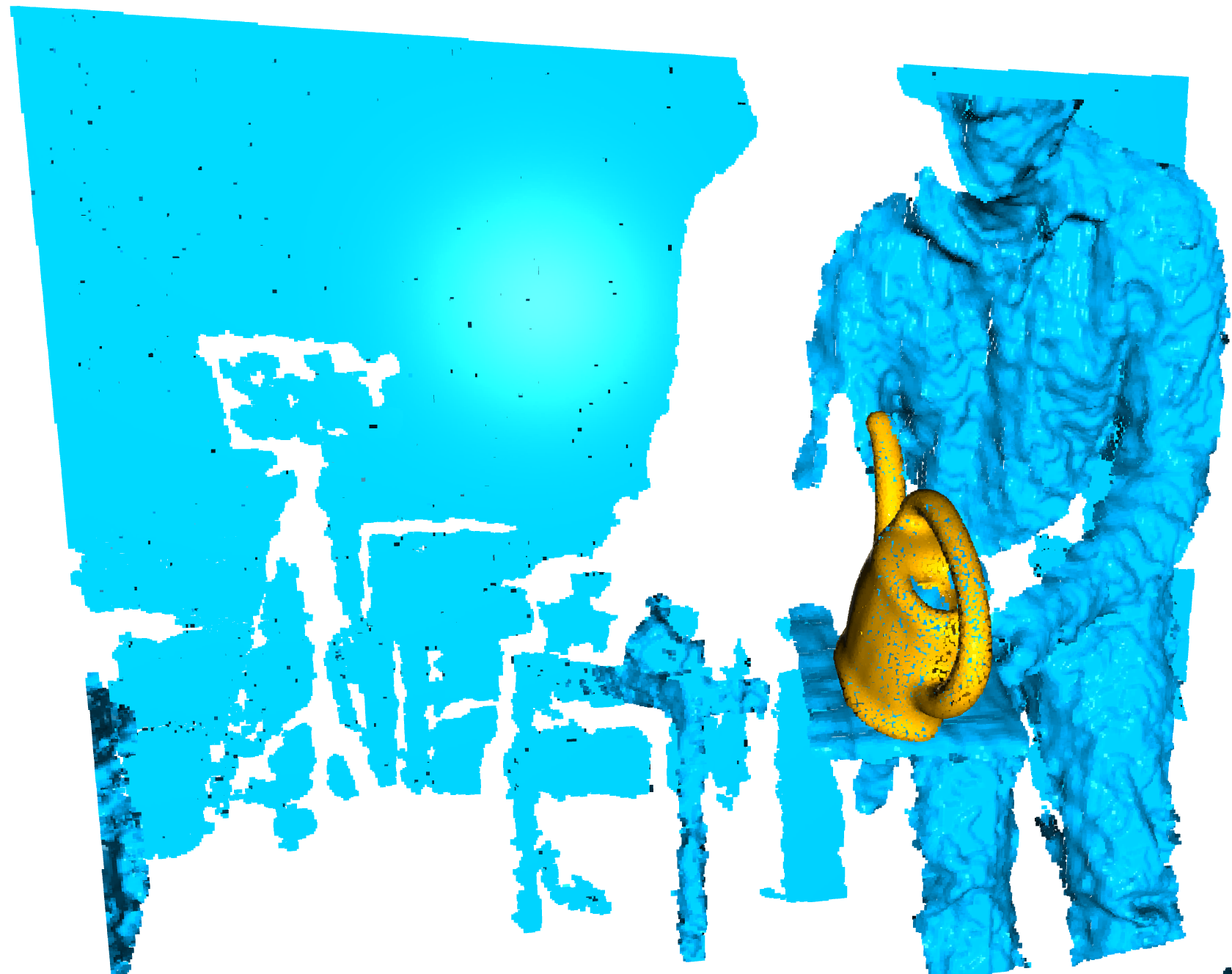}
    \caption{\label{real_scene} Qualitative results on the TUD-L dataset. (Left) Mask produced by our instance segmentation module. (Right) Object model registered to the input depth map using the pose predicted by Ours-OT.}
\end{figure}

%% file: fig/mAP_real.tex
\begin{figure}[!t]
    \centering
    \includegraphics[width=8.2cm]{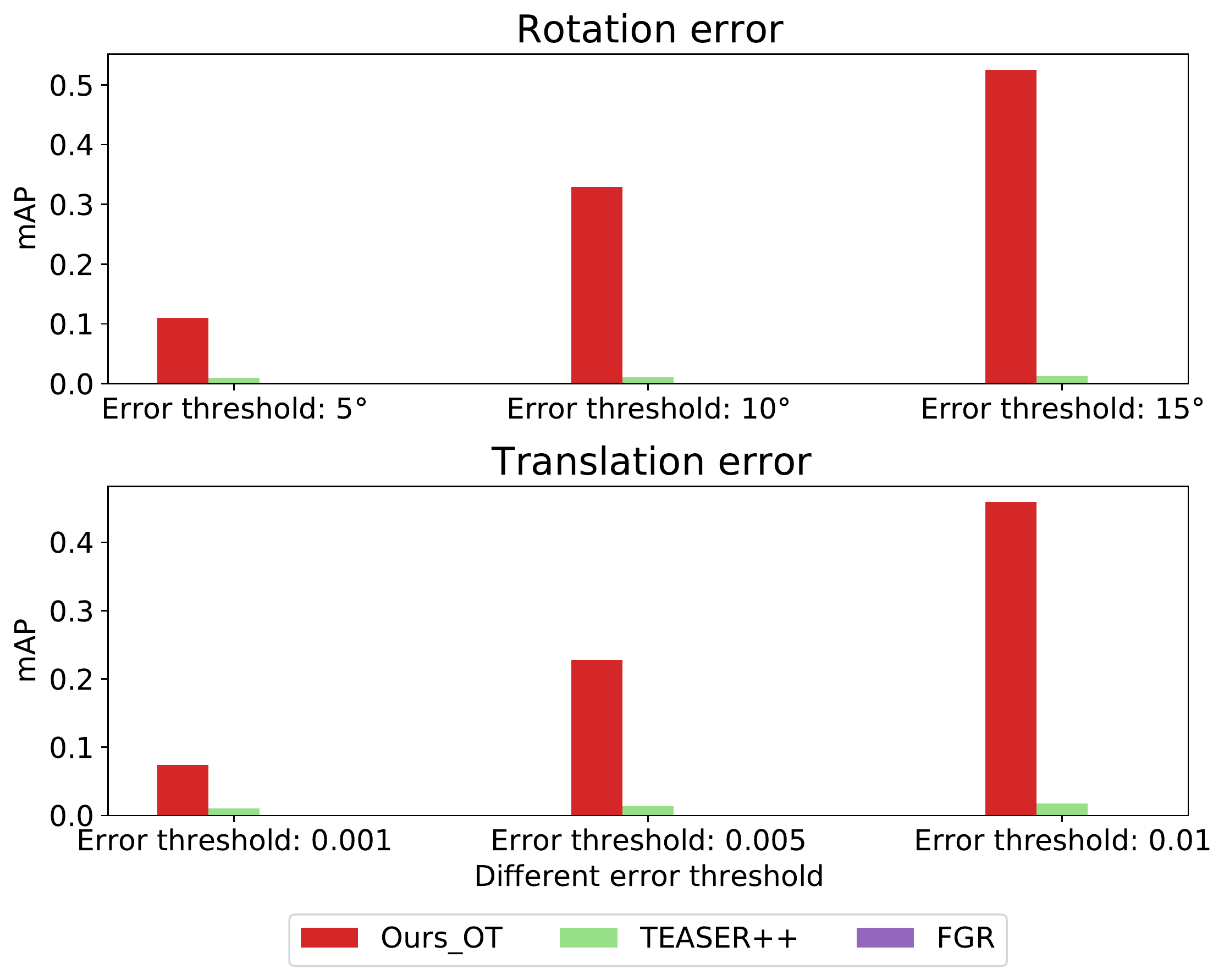}
    \setlength{\belowcaptionskip}{5pt}
    \caption{\label{mAP_real} Rotation and translation mAP for different thresholds on the TUD-L dataset.}
\end{figure}

%% file: table/inference_time.tex
\begin{table}[t]
    \centering
    \setlength{\tabcolsep}{6.5mm}{
    \begin{tabular}{@{}lccc@{}}
    \toprule
    \textbf{Methods}      &\textbf{512}  &\textbf{1024}  &\textbf{2048}  \\ \midrule
    \textbf{ICP}          & 0.018        & 0.028         & 0.054         \\
    \textbf{FGR}          & 0.032        & 0.059         & 0.132         \\
    \textbf{TEASER++}     & 0.025        & 0.085         & 0.249         \\
    \textbf{DCP}          & 0.013        & 0.018         & 0.053         \\
    \textbf{PRNet}        & 0.051        & 0.075         & 0.141         \\
    \textbf{Ours-OT}      & 0.033        & 0.052         & 0.112         \\ \bottomrule
    \end{tabular}}
    \caption{\label{tab:inference_time} Inference time (in seconds).}
\end{table}

%% file: tex/conclusion.tex
\section{Conclusion}
We have introduced the first learning-based approach to estimating the 6D pose of a self-occluded object immersed in a scene. Our method is intuitive, relying on an instance segmentation module followed by a pose estimation one. As evidenced by our empirical results, this strategy is effective, outperforming the state-of-the-art traditional and learning-based methods, which so far have mostly focused on registering partial views of the same object or scene. We believe that the problem we have tackled here is central in many practical applications. In particular, in the future, we will aim to deploy our approach for LiDAR-based registration in the context of non-cooperative rendezvous in space.

%does not require a complicated training strategy, yet yields more accurate results than other approaches that do. Furthermore, it has a reasonably low memory footprint and generalizes well to unseen data, making it broadly applicable. In the future, we will aim to deploy it for LiDAR-based registration, with a particular focus on the task of removing debris in low Earth orbit.